\documentclass{svproc}

\usepackage{url,babel}
\usepackage[numbers,sort&compress,sectionbib]{natbib}
\usepackage{amsmath,amssymb,enumitem}
\usepackage{graphicx, wrapfig}
\usepackage{textcomp}
\usepackage{xcolor, comment, setspace,empheq}
\usepackage[linesnumbered,ruled,vlined]{algorithm2e}
\usepackage[nottoc,notlof,notlot]{tocbibind} 

\usepackage{sectsty}

\newcommand{\f}{\mathbf{f}}
\newcommand{\R}{\mathcal{R}}
\newtheorem{defn}{Definition}
\newtheorem{cla}{Claim}

\makeatletter 
\renewcommand\@biblabel[1]{#1.} 
\makeatother

\makeatletter

\begin{document}

\mainmatter 
\title{Construction of Differentially Private Empirical Distributions from a low-order Marginals Set through Solving Linear Equations with $l_2$ Regularization}

\titlerunning{CIPHER} 
\author{Evercita C. Eugenio\inst{1} \and Fang Liu\inst{2}}
\tocauthor{Fang Liu and Evercita C. Eugenio}
\institute{Sandia National Laboratories, Livermore, CA 94550, USA\\
\email{eceugen@sandia.gov}
\and
University of Notre Dame, Notre Dame, IN 46556, USA\\
\email{fang.liu.131@nd.edu}}

\footnotetext[1]{The research was funded by the US National Science Foundation Awards \#1546373 and \#1717417. The publication has been
assigned the Sandia National Laboratories identifier SAND2021-0088 C.}

\maketitle

\begin{abstract}
We introduce a new algorithm, \textbf{C}onstruction of d\textbf{I}fferentially \textbf{P}rivate Empirical Distributions from a low-order marginals set t\textbf{H}rough solving linear \textbf{E}quations with $l_2$ \textbf{R}egularization (CIPHER), that produces differentially private empirical joint distributions from a set of low-order marginals. CIPHER is conceptually simple and requires no more than decomposing joint probabilities via basic probability rules to construct a linear equation set and subsequently solving the equations. Compared to the full-dimensional histogram (FDH) sanitization, CIPHER has drastic\-ally lower requirements on computational storage and memory, which is practically attractive especially considering that the high-order signals preserved by the FDH sanitization are likely just sample randomness and rarely of interest. Our experiments demonstrate that CIPHER outperfor\-ms the multiplicative weighting exponential mechanism in preserving original information and has similar or superior cost-normalized utility to FDH sanitization at the same privacy budget.
\keywords{differentially private empirical distributions and synthetic data, sign and statistical significance (SSS), full-dimensional histogram low-order marginals, computational storage and memory}
\end{abstract}

\section{Introduction}\label{sec:intro}
\subsection{Background and Motivation}
When releasing data sets for research and public use, protection of individual private information while still maintaining good utility of the data is of extreme importance. Even if a data set is de-identified, a data intruder may still be able to identify subjects by linking to other publicly available information \cite{narayanan2006how, narayanan2008robust, sweeny2013, tockar2014, culnane2017australia}. The intensified concerns on privacy call for more rigorous and mathematically sound privacy protection concepts and frameworks when sharing information. Differential privacy (DP) \cite{dwork2006calibrating, dwork2008differential} has emerged as one of the most powerful concepts to achieve that goal. DP provides rigorous mathematical guarantee for privacy protection without making strong assumptions about the intruder's background knowledge. One of the applications of DP is to generate differentially private distributions and individual-level synthetic or surrogate data so that data users may perform analysis on their own as if they had the original data. A simple way to sanitize data with minimal distributional assumptions on the original data is to sanitize the full-dimensional histogram (FDH), an empirical estimator of the joint distribution of all the attributes in the data. The approach is simple but has drawbacks when the data are multi-dimensional. First, there are likely a lot of empty cells in the histogram when its dimensionality $p$ is relatively large. Second, the high-order interactions among the attributes implicitly preserved by the FDH hardly represent any meaningful population-level signals. Lastly, it can be computationally costly to store or sanitize the FDH for a large $p$.

\subsection{Our Contributions}
\vspace{-3pt}
We propose a novel procedure, namely, \textbf{C}onstruction of d\textbf{I}fferentially \textbf{P}rivate Empirical Distributions from a low-order marginals set t\textbf{H}rough solving linear \textbf{E}quations with $l_2$ \textbf{R}egularization (CIPHER), to generate differentially private empirical distributions from which individual-level data surrogate or synthetic data can be easily obtained. CIPHER is based on the general knowledge that the population-level signals in real-life data are oftentimes contained in a set of low-order marginals. The advantages of CIPHER and its practical significance's are summarized as follow.
\begin{itemize}
\item CIPHER is conceptually simple and requires nothing than decomposing joint probabilities among attributes via basic probability rules to construct and solve a linear equation set. It does not impose strong assumptions on the local data. The computational cost for solving the equation set is expected to be low once the equation set is constructed.
\item CIPHER can automatically correct the inconsistency across the marginals of the same variable that appear in multiple histograms caused by differentially private sanitization, without explicitly incorporating constraints. 
\item Compared to the FDH sanitization, the set of low-order marginals that CIPHER employs has drastically lower requirements for computer storage and memory. For example, compared to 9,765,625 cells resultant from the FDH of 10 attributes with 5 levels each, there is a 95.4\% and 99.99\% reduction in the number of cells -- 62,200 and 8,440, respectively -- if CIPHER uses the set of four-way histograms (210) and the set of two-way histograms (45), respectively.
\item If a data user is provided with a set of differentially private low-order margin\-als with inconsistent counts due to sanitization, she may apply CIPHER to generate individual-level data to meet her analysis needs, without incurring additional privacy costs.
\end{itemize}

\subsection{Related Work}\label{sec:work}
\vspace{-3pt}
There exist some methods that generate differentially private empirical distribu\-tions or synthetic data from a set of low-order statistics, even though they might be proposed originally for different purposes. Each approach has its own pros and cons, and differs from CIPHER in either the formulation of the query sets, or computationally, or methodologically. 

\citet{barak2007privacy} propose a method based on Fourier transforms. The linear programming employed by the method is a bottleneck for this algorithm especially for large $p$. \citet{chen2015differentially} form differentially private histograms from attribute clusters. The formation of optimal attribute clusters is an NP-hard problem and the authors introduce an approximation algorithm that does not guarantee optimality. Compared to these two methods, CIPHER is less costly computation-ally, though it requires a careful layout of the equation set from which the sanitized empirical distribution is calculated. \citet{liu2016model} proposes a model-based approach  to generate differentially private synthetic data (modips) in the Bayesian framework. The modips has practical limitations in terms of both model construc-tion as well as sufficient statistics sanitization for certain data types or large $p$. \citet{machanavajjhala2008privacy} demonstrate that the Multinomial-Dirichlet model sanitization leads to poor inferences due to data sparsity when it is applied to release the commuting patterns of the US population data. 
\citet{bowen2016comparative} also show that the approach has worse performance than the FDH sanitization via the Laplace mechanism and the modips approach at the same privacy budget. CIPHER, compared to the modips and Multinomial-Dirichlet model methods, does not require specification of a statistical model on the original data. 
PrivBayes \cite{Zhang2014} has some similarity with CIPHER in the sense that both rely on a reduced set of relationships among the attributes to capture the signals in the data. On the other hand, PrivBayes is different from CIPHER in that it explores the conditional independence among the attributes to approximate the joint distribution of the attributes via a directed acyclic graph -- Bayesian network, whereas CIPHER does not need data-driven selection of the set of queries or formulation of a model, and can save all privacy budget toward the sanitization process. The price paid by CIPHER  is that the set of queries it generates the empirical distribution from might not be the most representative of the underlying population signals and relations among the attributes compared to the Bayesian network selected by PrivBayes by spending a certain amount of privacy cost.  
\citet{hardt2012simple} propose the iterative Multiplicative Weights via Exponential Mechanism (MWEM) approach. The MWEM algorithm achieves a near optimal bound on the $l_{\infty}$ error for the queries used to generate synthetic samples if the number of iterations $T$ is optimized. However, it can be very challenging to choose the optimal $T$ and the accuracy of MWEM is highly dependent on $T$. MWEM is an iterative procedure and each iteration incurs privacy cost due to it accessing the original data to fetch the query selected by the Exponential mechanism, which is subsequently sanitized by the Laplace mechanism. CIPHER is non-iterative and has different privacy cost than MWEM. We will discuss more on the differences between CIPHER and MWEM, and between CIPHER and PrivBayes in Section \ref{sec:cipher} after presenting the detailed steps of the CIPHER algorithm.

The rest of the paper is organized as follows. Section \ref{sec:method} introduces the CIPHER procedure. Section \ref{sec:experiment} compares the CIPHER with several other sanitization methods on the statistical utility of the synthetic data in simulated and real-life data. It also proposes the SSS (Sign and Statistical Significance) assessment to evaluate the inferences based on differentially private synthetic data against the original inferences. Section \ref{sec:discussion} provides some concluding remarks and discusses future research directions.

\section{CIPHER}\label{sec:method}
\subsection{Preliminaries}\label{sec:prelim}
\vspace{-3pt}
Consider a data set $D$ and a query or a set of queries $\f$ about $D$. DP provides a rigorous mathematical conceptual framework to protect individual privacy information when releasing the query results to $\f$. 
\vspace{-3pt}
\begin{defn}[\textbf{$\epsilon$-differential privacy} \cite{dwork2006calibrating}]\label{defn:dp} A randomized mechanism $\R$ satisfies $\epsilon$-differential privacy if for all data sets $D_1$ and $D_2$ differing on one individual and all result subsets $S$ to query $\f$, $e^{-\epsilon} \leq \frac{\Pr [ \R \left( \f(D_1) \right) \in S ]}{\Pr [\R \left( \f(D_2) \right) \in S ]} \leq e^{\epsilon}$ for $\epsilon>0$.
\end{defn}
\vspace{3pt}
\noindent The formulation of privacy via the DP is robust and guards against the worst-case scenario as it does not impose any assumptions about the behavior or the background knowledge of data intruders. $D_1$ and $D_2$ differing by one individual can be interpreted in two ways: $D_1$ is one individual more or less than $D_2$, or $D_1$ and $D_2$ are of the same size but have difference in attributes values in exactly one individual. $\epsilon$ is often referred as the privacy budget and is pre-specified. The smaller $\epsilon$ is, the more privacy protection is imposed on the individuals in the data. Interested readers may refer \citet{bowen2016comparative} and \citet{liu2021} for brief discussions on the choice of $\epsilon$. Besides the pure $\epsilon$-DP in Definition \ref{defn:dp}, there are several relaxed versions, such as the $(\epsilon, \delta)$-DP and $(\epsilon, \delta)$-probabilistic DP. The former relaxes the bounds on the ratio of $\Pr(\R(\f(D_1)\in S)/ \Pr(\R(\f(D_2))\in S))$, whereas the latter bounds the probability the pure $\epsilon$-DP is violated. In both cases, setting $\delta$ at 0 reduces to  the pure $\epsilon$-DP.

In practice, a data set can be  queried  multiple times. Every time a query result is released, some privacy is lost for the individuals in the data set.  The sequential composition and parallel composition principles presented in Definition \ref{defn:sequential} are useful for tracking and counting of privacy costs when designing differentially private mechanisms and releasing query results form a data set. 
\begin{defn}[\textbf{sequential composition and parallel composition} \cite{mcsherry2009privacy}]\label{defn:sequential} \hspace{6pt}\\
Let $f_i$ for $i=1,...,K$ represent a set of queries on data $D$ and $\epsilon$ be the total privacy budget. Denote by 
$\R_i$ a randomization mechanism of $\epsilon_i$-DP. The sequential composition states that the sequence of $\R_i(D)$ provides $(\sum_i \epsilon_i)$-DP. The parallel composition states that the sequence of $\R_i(D_i)$ provides $\max_i{\epsilon_i}$-DP, where \\ $\{D\}_{i=1,...,K}$ are arbitrary disjoint subsets of $D$.
\end{defn}

There are a variety of mechanisms to provide differentially private results. Interested readers may refer to \citet{dwork2014algorithmic} for some of the commonly used DP mechanisms. Here we mention the Laplace mechanism, that is used in the experiments in Section \ref{sec:experiment}. 
\vspace{-3pt}\begin{defn}[\textbf{Laplace mechanism} \cite{dwork2006calibrating}]\label{defn:laplace}
The Laplace mechanism of $\epsilon$-DP \\generates sanitized query results as in $\f^{\ast}(D)\sim \mbox{Lap}(\f(D),\Delta_{\f} / \epsilon)$, where $\Delta_{\f} = \\ {\max}_{D_1, D_2} \| \f(D_1)-\f(D_2) \|_1$ is the $l_1$ global sensitivity of query $\f$, for all data sets $D_1$, $D_2$ differing by one element. 
\end{defn}
The larger $\Delta_\f$ is, the more noise would be injected to $\f$ to satisfy the $\epsilon$-DP. Generalization of the Laplace mechanism include the  Gaussian  mechanism and Generalized Gaussian  mechanism that is built upon the $l_p$ norm ($p\ge1$) \citep{dwork2014algorithmic, liu2016generalized}, among others. Definition  \ref{defn:exponential} presents the exponential mechanism, which is used in the MWEM procedure implemented in the experiments in Section \ref{sec:experiment}.
\begin{defn}[\textbf{exponential mechanism} \citep{mcsherry2007mechanism}]\label{defn:exponential}
Let $u$ be a utility function that assigns a score to each possible output of a query to data $D$. The Exponential mechanism that satisfies $\epsilon$-DP releases query result $f^{\ast}(D)$ with probability  
$$\textstyle\exp ( u(f^{\ast}(D);D) \frac{\epsilon}{2\delta_u} )/\int u(f^{\ast}(D); D) \frac{\epsilon}{2\delta_u} d(f^{\ast}(D)),$$ where $\delta_u$ is the maximum change in score $u$ with one element change in data $D$.
\end{defn}

\subsection{The CIPHER Procedure}\label{sec:cipher}
We propose the CIPHER procedure to generate differentially private empirical distributions and individual-level synthetic data from a set of low-order marginals. As mentioned in Section\ref{sec:intro}, the main motivation for CIPHER is to reduce the dimension of the query set from which individual-level data can be generated to save on storage and memory, while preserving the population-level signals in the original data.  
\vspace{-12pt} \begin{figure}[!htb]
\centering
\includegraphics[scale=0.52]{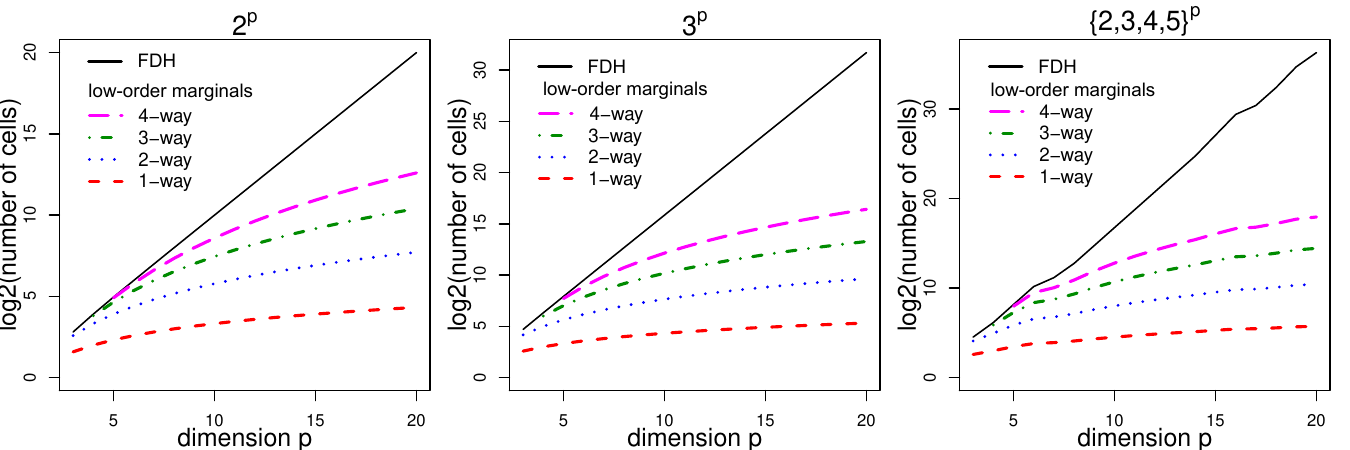}
\vspace{-12pt}
\caption{log(number of stored cell in low-order marginals) vs the number of attributes $p$ (leftmost: each attribute is binary; middle: each attributes has 3 levels; rightmost: the number of levels ranges from 2 to 5 among the $p$ attributes)} \label{fig:storage} 
\end{figure} \vspace{-12pt}
Figure \ref{fig:storage} shows the drastic reduction in the number of cells that need to be stored if the sets of 1-way, 2-way, 3-way, and 4-way low-order marginals are used in place of the full table for varying $p$ (the number of attributes in the original data). The degree of the low-order marginals from which higher-order marginals are generated is allowed to grow with $p$; but it should always be kept in mind that interactions of very high orders are rarely of interest in real life and are also hard to explain. 

The CIPHER algorithm is presented in Algorithm \ref{alg:CIPHER}. In brief, the algorithm starts with a set of low-order marginals $\mathcal{Q}$ and arrives at a solution of the differentially private empirical distribution via a stepwise but non-iterative fashion, without involving complex sampling algorithms. The low-order marginals in $\mathcal{Q}$, which do not have to be of the same dimension, capture the important signals and relationships among the attributes in the original data. Two special cases of

\begin{algorithm}[H]
\caption{CIPHER} \label{alg:CIPHER}
\small
\SetAlgoLined
\SetKwInOut{Input}{input}
\SetKwInOut{Output}{output}
\Input{original data $D$ ($n\times p$); query set $\mathcal{Q}$; privacy budget $\epsilon$; number of empirical distributions or synthetic data sets $m$; $l_2$ regularization constant $\lambda$.}
\Output{differentially private empirical distributions $P(\mathbf{X})^{(l)}$ or data sets $\tilde{D}^{(l)}$ for $l=1,\ldots,m$.}
Denote the lowest dimension of the marginals in $\mathcal{Q}$ by $p_0$\;
\For{$l=1,\ldots,m$}{
Sanitize all queries $\in \mathcal{Q}$ via a mechanism of $\epsilon$-DP (e.g., $\tilde{q}^{(l)}_i=q_i+$Lap$(0,\epsilon/(m|\mathcal{Q}|))$ for $i=1,\ldots,|\mathcal{Q}|$)\;
\For{$j=p_0+1,\ldots,p$}{
List all $j$-way marginals $\mathcal{T}_j$\;
\For{each $q_i\not\in(\mathcal{T}_{j+1}\cap\mathcal{Q})$}{
1) Denote the variables involved in query $q_i$ by $\mathcal{X}_i$ and let $p_i\!=\!|\mathcal{X}_i|$\;
2) Randomly pick a variable out of $\mathcal{X}_i$ and label it as $X_{i1}$, and the rest as $X_{i2},\ldots, X_{i,p_i}$. Denote the number of bins or levels of $X_{ik}$ by $K_{ik}$ for $k=1,\ldots,p_i$\;
3) \For{$k=2,\ldots,(p_i-1)$}{
Define $\mathbf{b}_k\!=\! P(X_{i1}\ne K_{i1}|\mathbf{X}_{i}\setminus (X_{i1},X_{ik})$ $=\textstyle\sum_{X_{ik}}P(X_{i1}\ne K_{i1},X_{ik}|\mathbf{X}_{i}\setminus (X_{i1},X_{ik}))=\mathbf{A_kz_k}=$
$\!\sum_{X_{ik}}\!\!P(X_{ik}|\mathbf{X}_{i}\!\setminus\!(X_{i1}, X_{ik})) P(X_{i0}\!\ne\!K_{i1}|\mathbf{X}_i\!\setminus\!(X_{i1}, X_{ik}), X_{ik})$, where $\mathbf{z}_k$ is the conditional probability of $(X_{i1}\ne K_{i1})$ given the rest of variables in $\mathcal{X}_i$, $\mathbf{A}_k$ is either observed or calculated from step $j-1$, and $(X_{i1}\ne K_{i1})$ represents the vector $(X_{i1}=1,\ldots,X_{i1}\!=\!K_{i1}-1)$.\;}
4) Let $\mathbf{b}=(\mathbf{b}_1,\ldots, \mathbf{b}_{p_i-1})^T$, 
$\mathbf{z}=(\mathbf{z}_1,\ldots, \mathbf{b}_{p_i-1})^T$, and $\mathbf{A}=$ Diag$\{\mathbf{A}_1,\ldots,\mathbf{A}_{p_i-1}\}$; solve for $\mathbf{z}$ from $\mathbf{Az}=\mathbf{b}$ as in $\mathbf{z}=(\mathbf{A^TA+\lambda I})^{-1}\mathbf{A^Tb}$, where $I$ is the identity matrix\;
5) Calculate the private empirical joint probability the variables in $\mathcal{X}_i$: $P(\mathcal{X}_i)\!=\!\mathbf{z}\cdot P(\mathbf{X}_i\setminus X_{i1})$\;
}
}
Correct negativity and normalize the private empirical joint probability $P(\mathbf{X})^{(l)}=P(X_1,...,X_p)^{(l)}$, and generate private data $\tilde{D}^{(l)}$ of size $n$ from $P(\mathbf{X})^{(l)}$ if needed\;
}
\end{algorithm} 
\noindent $\mathcal{Q}$ are the single $p$-way full table and the set of $p$ one-way contingency tables, respectively. Forming $\mathcal{Q}$ can be guided by the domain knowledge so not to consume the additional privacy in the original data. If the domain knowledge is not available or the data curator prefers to choose $\mathcal{Q}$ using the information of the original data, the total privacy budget will need to be divided between the selection of $\mathcal{Q}$ and the CIPHER algorithm itself. In the rest of the discussion, we assume $\mathcal{Q}$ is preset before the application of the CIPHER algorithm. 
 
\begin{cla}\label{rem:dp}
The CIPHER algorithm satisfies $\epsilon$-DP.
\end{cla}
\noindent The satisfaction of DP in CIPHER is straightforward to establish. The only time at which the original data are probed during the application of CIPHER is when the queries in $\mathcal{Q}$ are sanitized. All together, the data are accessed $mK$ times with a privacy budget of $\epsilon/(mK)$ per access. Per the sequential composition, the total privacy budget for releasing the privacy-preserving empirical distribution is maintained at $(mK)\epsilon/(mK)=\epsilon$.

We recommend setting the number of sanitized distribution sets $m$ in the algorithm at a small number, say 1 to 5. $m>1$ is specified when there is a need to account for the randomness and uncertainty in the released information due the sanitization and synthesis. In addition, as long as $m$ is not too large so that the total privacy budget is not spread too thin over the multiple sets (each synthetic set receives $1/m$ of the total privacy budget per the sequential composition theorem), the precision gained by averaging over $m$ sets of synthetic data could outweigh the additional noises introduced from releasing multiple sets than a single set. In the case of statistical inferences based on $m$ sets of synthetic data are of interest, they can be obtained through the inferential combination rules in \cite{liu2016model}.

The reason for using the $l_2$ regularization (aka the Tikhonov regularization) to solve for $\mathbf{z}$ from $\mathbf{Az}=\mathbf{b}$ is that the columns of $\mathbf{A}$ are linearly dependent and $A^TA$ is not of full rank. The $l_2$ regularization is known for solving ill-posed problems like $\mathbf{Az}=\mathbf{b}$ when the solution $\mathbf{z}$ is not unique due to the singularity of $\mathbf{A}$ \citep{tikhonov1963solution,tikhonov2013numerical}. 
It adds a small positive constant $\lambda$ to $\mathbf{A^TA}$ and calculates $\mathbf{z}=(\mathbf{A^TA +\lambda I})^{-1}\mathbf{A^Tb}$. Since $\mathbf{A}$ is block-diagonal, taking the inverse of $\mathbf{A^TA+\lambda I}$ is computationally cheap, relatively speaking, even if the linear equation set is large. Regarding the choice of hyper-parameter $\lambda$, our empirical studies suggest that the solutions are relatively insensitive to $\lambda$ in the case of CIPHER as long as $\lambda$ is relatively small ($o(1)$). 

If one or more attributes end up appearing in $\ge2$ marginals in $\mathcal{Q}$, then after the sanitization, the empirical distributions of the shared attributes would be inconsistent across different low-order marginals. The good news that the inconsistency is automatically averaged out for CIPHER when solving the linear equation set, without a need for ad-hoc step to correct for the inconsistency, because the formulation of the linear equation set implicitly builds in the constrains. This offers a great advantage over the methods that employ ad-hoc correction procedures (e.g.,\cite{barak2007privacy, hay2010boosting}). 

The cell probabilities in the low-order marginals in $\mathcal{Q}$ after DP sanitization can be $<0$ or $\ge1$, so are the solutions for the conditional probabilities in some intermediate steps of the CIPHER algorithm. We suggest performing a one-time correction in the last step of generating the empirical distribution. 

Both CIPHER and MWEM may use a pre-specified set of linear queries to generate differentially private  empirical distributions, but they differ methodolo-gically and algorithmically. First, MWEM relies on an iterative multiplicative weighting procedure whereas CIPHER is not iterative. Second, the queries in CIPHER are sanitized one time through a DP mechanism (say the Laplace sanitizer) before being fed to the algorithm. By contrast, each iteration in the MWEM algorithm incurs a privacy cost due to it accessing the original data to fetch the query selected by the exponential mechanism, which is subsequently sanitized by the Laplace mechanism. As a result, the two algorithms spend different amounts of privacy cost per query for a given total privacy budget. Suppose the total budget is $\epsilon$, and the number of queries in $\mathcal{Q}$ is denoted by $|\mathcal{Q}|$. If we use equal allocation of the privacy budget, then each query in $\mathcal{Q}$ gets a budget of $\epsilon/|\mathcal{Q}|$ in the CIPHER algorithm. The sanitization of each query selected by the exponential mechanism costs $\epsilon/(2T)$ in the MWEM algorithm. On the other hand, a query can be selected multiple times throughout the $T$ iterations. Let $c_k$ denote the number of times that $q_k\in\mathcal{Q} $ is selected among the $T$ iterations. Note $\sum_{k=1}^{|\mathcal{Q}|}{c_k}=T$. Unless $c_k/(2T) > |\mathcal{Q}|^{-1}$ or $c_k/\sum_{k=1}^{|\mathcal{Q}|}{c_k} > 2|\mathcal{Q}|^{-1}$, then the budget allocated to query $q_k$ in the MWEM algorithm would always be smaller than that in CIPHER. In other words, the selection probability for a query needs to at least double the average selection probability ($1/|\mathcal{Q}|$) in order to that query to receive more privacy budget in MWEW than the amount of budget it receives in CIPHER. In addition, our own experiences from running the MWEM algorithm suggest that choosing the ``right'' number of iterations $T$ for MWEM can be challenging. $T$ too small is not sufficient to allow the empirical distribution to fully capture the signals summarized in the queries; and $T$ too large would lead to a large amount of noises being injected as the privacy budget has to be distributed across the $T$ iterations, eventually leading to a useless synthetic data set as each iteration costs privacy. 
PrivBayes offers similar benefits on data storage and memory, similar to CIPHER, given that PrivBayes is built in the framework of Bayesian network that is is known for its ability of saving considerable amounts of memory over full-dimensional tables if the dependencies in the joint distribution are sparse. On the other hand, PriBayes starts with model building that costs privacy budget. It is also well known approximate structure learning of a Bayesian network is NP-complete. In addition, Bayesian networks would force attributes in a data set to be in a causal relationship. Finally, PrivBayes proposes a surrogate function for mutual information, on which the quality of the released data replies, requires some effort for efficient  computation. In comparison, the underlying analytical and computational techniques for CIPHER are standard and require nothing than joint  probability decomposition and solving linear equations. 

\subsection{Example: CIPHER for the 3-variable Case} 
We illustrate the CIPHER procedure with a simple example. Say the original data contain 3 variables $(p=3)$. Denote the 3 variables by $V_1,V_2,V_3$ with $K_1,K_2$ and $K_3$ levels, respectively. Let $\mathcal{Q}=\{\mathcal{T}(V_1,V_2),\mathcal{T}(V_2,V_3),\mathcal{T}(V_1,V_3)\}$ that contains all the 2-way contingency tables. Therefore, $p_0=2$ in Algorithm \ref{alg:CIPHER}. WLOG, suppose $V_3$ is $X_0$ in Algorithm \ref{alg:CIPHER}. We first write down the relationships among the probabilities, which are
$$
\begin{cases}
\Pr(V_3|V_1)=\textstyle\sum_{V_2}\Pr(V_3,V_2|V_1)=\sum_{V_2}\Pr(V_3|
V_1,V_2)\Pr(V_2|V_1)\\ \Pr(V_3|V_2)=\textstyle\sum_{V_1}\Pr(V_3,V_1|V_2)=\sum_{V_1}\Pr(V_3|
V_1,V_2)\Pr(V_1|V_2)
\end{cases}.$$
We now convert the above relationships into the equation set $\mathbf{b=Az}$. Specifically, $\mathbf{b}=(\Pr(V_3| V_1)\setminus\Pr(V_3=K_3| V_1)$, and $\Pr(V_3| V_2)\setminus\Pr(V_3=K_3| V_1))^T$ is a known vector of dimension $(K_1+K_2)(K_3-1)$, $\mathbf{z}= \Pr(V_3| V_1, V_2)\setminus\Pr(V_3=K_3| V_1,V_2)$ is of dimension $K_1K_2(K_3-1)$, $\mathbf{A}$ is a known diagonal matrix with $K_3-1$ identical blocks, and each block is a $(K_1+K_2)\times(K_1K_2)$ matrix comprising the coefficients (i.e., $\Pr(V_1|V_2), \Pr(V_2|V_1)$ or 0) associated with $\mathbf{z}$. After $\mathbf{z}$ is solved from $\mathbf{b=Az}$, the joint distribution of $\Pr(V_1,V_2,V_3)$ is calculated by $\mathbf{z}\cdot\Pr(V_1,V_2)$. The experiments in Section \ref{sec:experiment} contain more complicated applications of CIPHER.

\section{Experiments}\label{sec:experiment}
We run experiments with simulated and real-life data to evaluate CIPHER, and benchmark its performance against MWEM and the FDH sanitization in this paper. Both MWEM and FDH inspired our work, the former conceptually as stated in Section \ref{sec:cipher}, while the latter motivated us to develop a solution with decreased storage cost for sanitized queries. In addition, more complex algorithms are unlikely to beat a simpler and easier-to-deploy flat algorithm such as FDH, per conclusions from previous studies \cite{hay2016principled, bowen2016comparative} when $n$, $\epsilon$, or $p$ is large. In addition, both MWEM and FDH are straightforward to program and implement. Our goal is to demonstrate in the experiments that CIPHER performs better than MWEM in terms of the utility of sanitized empirical distributions and synthetic data, and delivers non-inferior performance compared to FDH with significant decreased storage costs.

When comparing the utility of synthetic data generated by different procedures, we not only examine the degree to which the original information is preserved on descriptive statistics such as mean and $l_q$ ($q>0$) distance, we also examine the information preservation in statistical inferences on population parameters. Toward that end, we propose the \emph{SSS assessment}. The first S refers to the Sign of a parameter estimate, and the second and third S' refer to the Statistical Significance of the estimate against the null value in hypothesis testing. Whether the sign and statistical significance in the estimate between the original and synthetic data are consistent leads to 7 possible scenarios (Table \ref{table:3S}). Between the best and the worst scenarios, there are 5 other possibilities. II+ and I+ indicate an increase in Type II (false negative) and Type I (false positive) error rates, respectively, from the original to the sanitized inferences, so do II- and I-, but the latter two also involve a sign change from the original to the sanitized inferences.
\vspace{-6pt}
\begin{table}[!htb]
\caption{Preservation of \textbf{S}igns and \textbf{S}tatistical \textbf{S}ignificance on an estimated parameter (the SSS assessment) }\label{table:3S}\vspace{-9pt}
\centering 
\resizebox{1\linewidth}{!}{
\begin{tabular}{l| cc| cc cccc}
\hline
parameter estimate & \multicolumn{2}{c|}{Best} & Neutral & II+ & I+ & II- & I- & Worst \\
\hline
matching \textbf{S}igns between non-private and sanitized? & Y & Y & N & Y & Y & N & N & N\\
non-private \textbf{S}tatistical \textbf{S}ignificance? & Y & N & N & Y & N & Y & N & Y \\
sanitized \textbf{S}tatistical \textbf{S}ignificance? & Y & N & N & N & Y & N & Y & Y \\
\hline
\end{tabular}}
\end{table}\vspace{-12pt}

\subsection{Experiment 1: Simulated Data}\label{sec:simulations}
The data were simulated via a sequence of multinomial logistic regression models with four categorical variables and two samples size scenarios at $n=200$ and $n=500$, respectively. For the FDH sanitization, there are 36 $(2\times2\times3\times3)$ cells in the 4-way marginals. For the CIPHER and MWEM algorithms, we consider 3 different query sets $\mathcal{Q}$: (1) $\mathcal{Q}_3$ contains all 3-way marginals, leading to 32 cells (88.9\% of the 4-way); (2) $\mathcal{Q}_2$ contains all six 2-way marginals, leading to 20 cells (55.6\% of the 4-way). Five privacy budget scenarios $\epsilon=(e^{-2},e^{-1},1,e^{1},e^{2})$ were examined. To account for the uncertainty of the sanitization and synthesis in the subsequent statistical inferences, $m=5$ synthetic data sets were generated. We run 1,000 repetitions for each $n$ and $\epsilon$ scenario to investigate the stability of each method. When examining the utility of the differentially private data, we present the cost-normalized metrics wherever it makes sense. The cost is defined as the number of cells used to generate a differentially private empirical distribution (36 for FDH; 32 for CIPHER 2-way and MWEM 2-way; and 20 for CIPHER 3-way and MWEM 3-way).

\begin{figure}[!htb]
\vspace{-6pt} \centering
\includegraphics[scale=0.205,trim={0.2cm 0.6cm 1cm 1.3cm},clip]{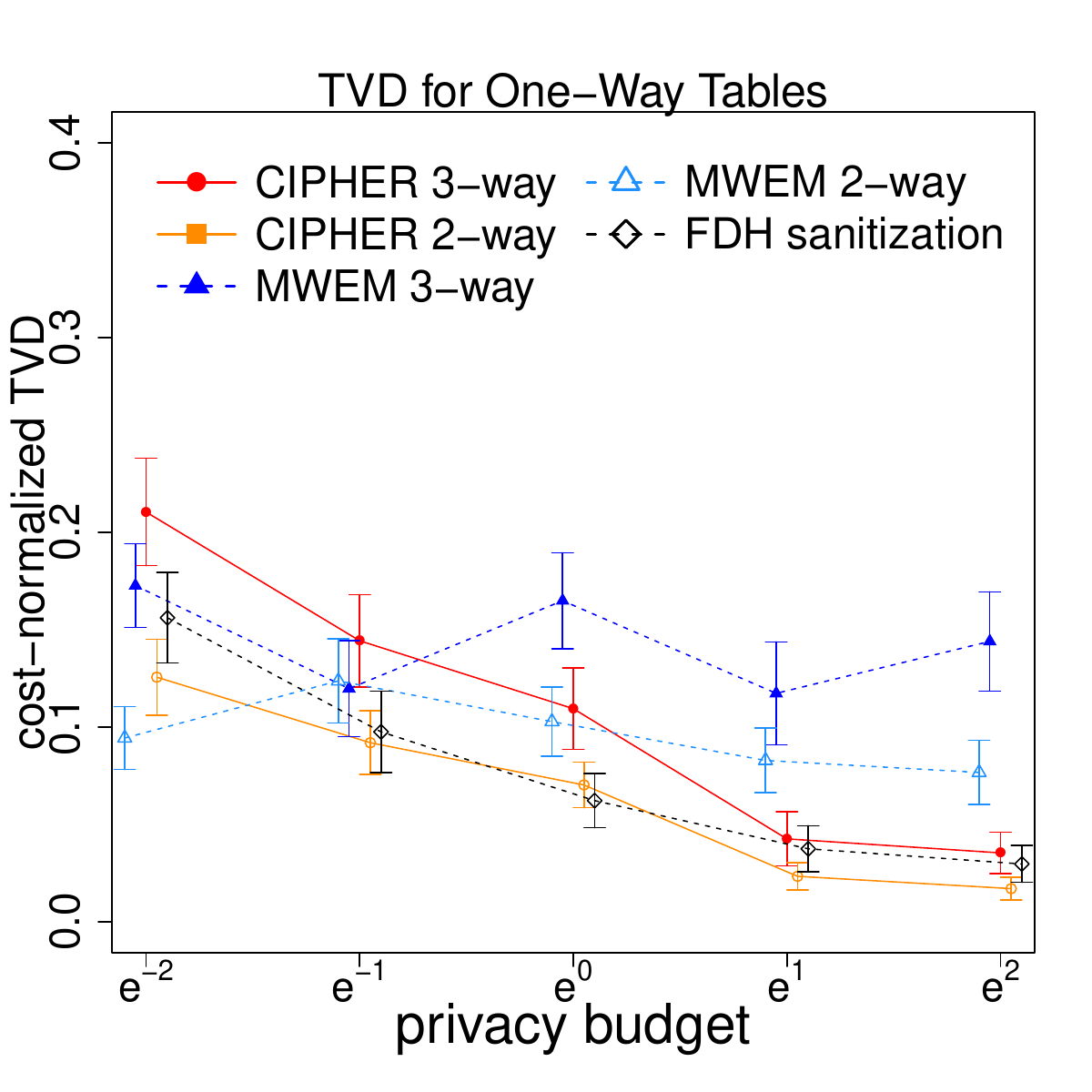}
\includegraphics[scale=0.205,trim={0.2cm 0.6cm 1cm 1.3cm},clip]{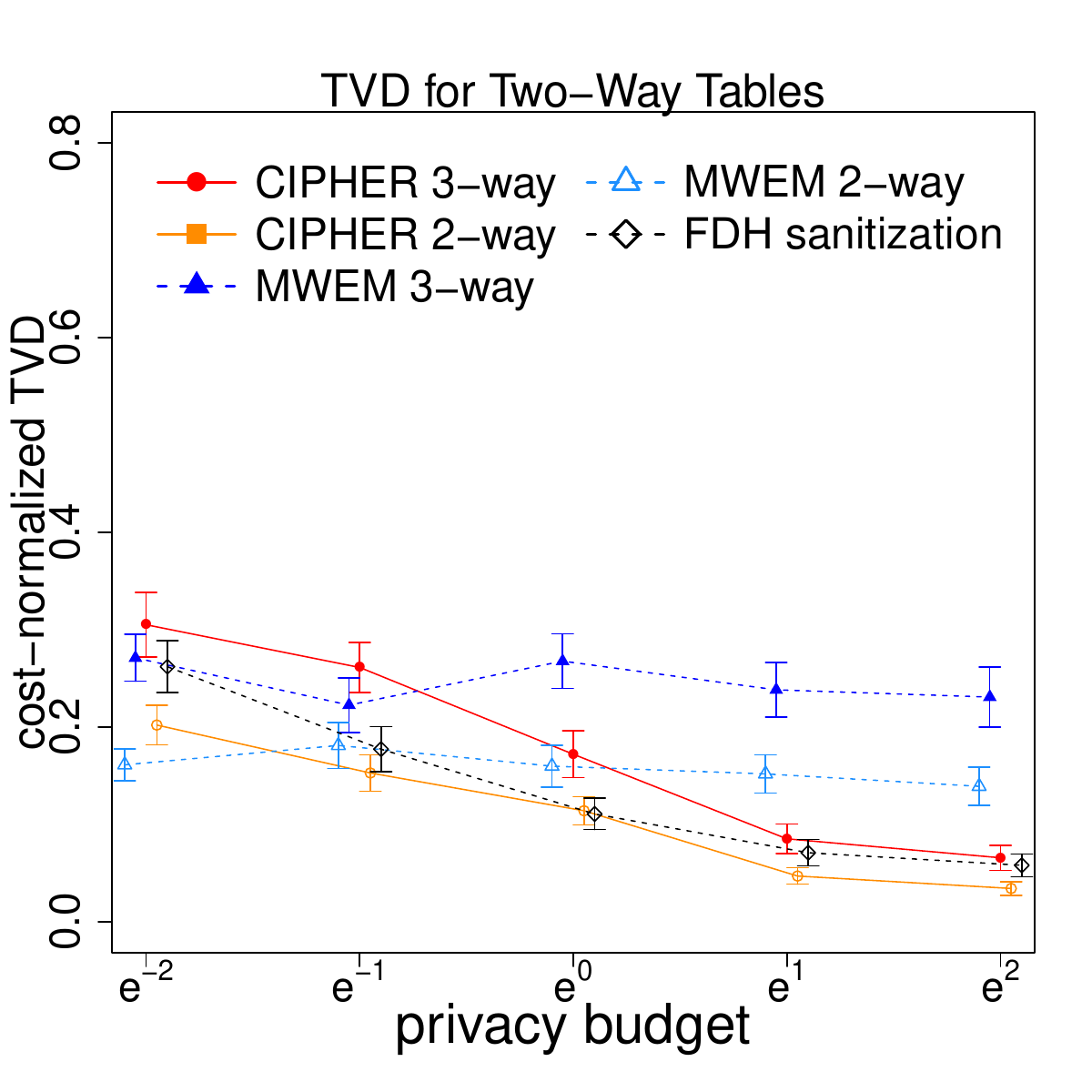}
\includegraphics[scale=0.205,trim={0.2cm 0.6cm 1cm 1.3cm},clip]{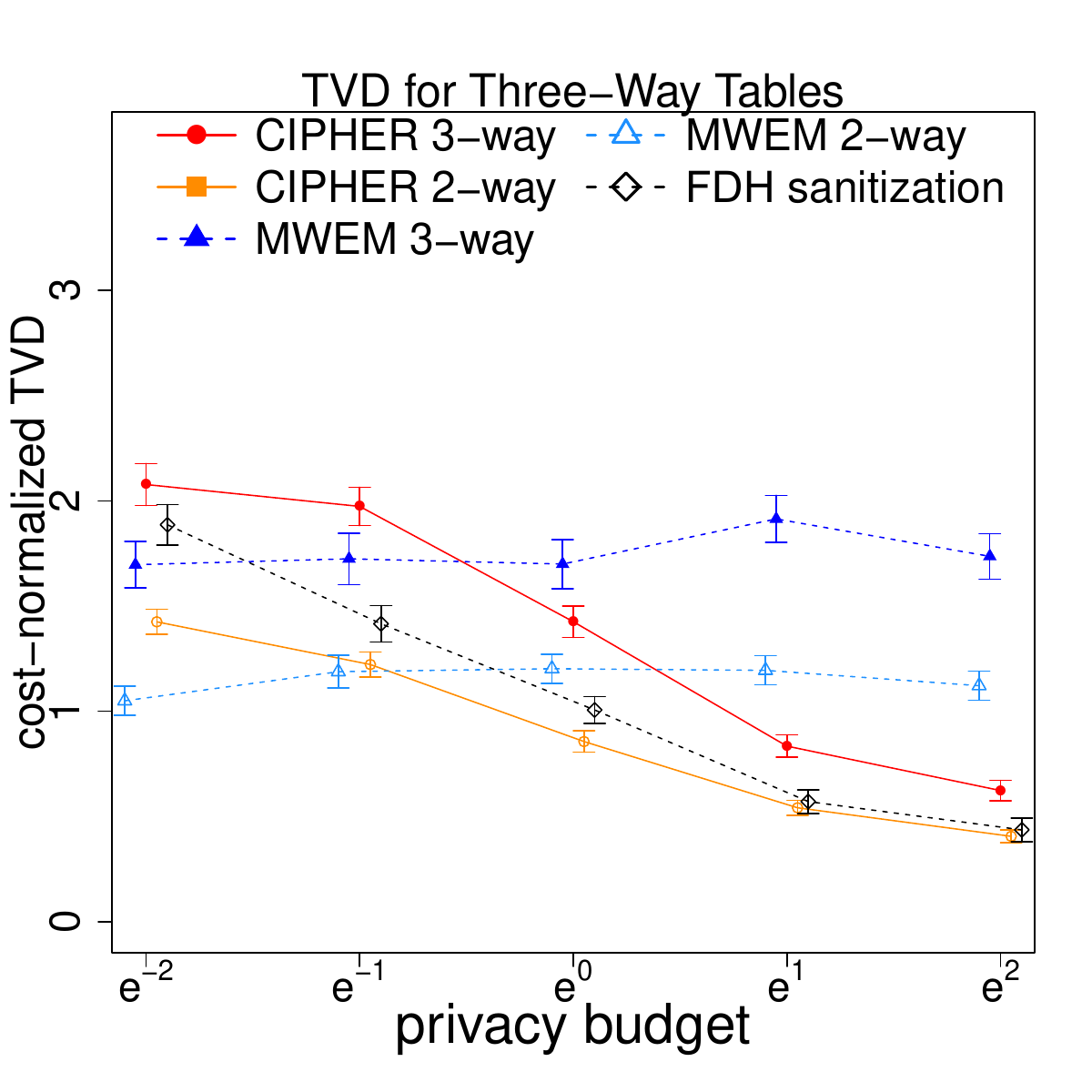}\\
\includegraphics[scale=0.205,trim={0.2cm 0.6cm 1cm 1.3cm},clip]{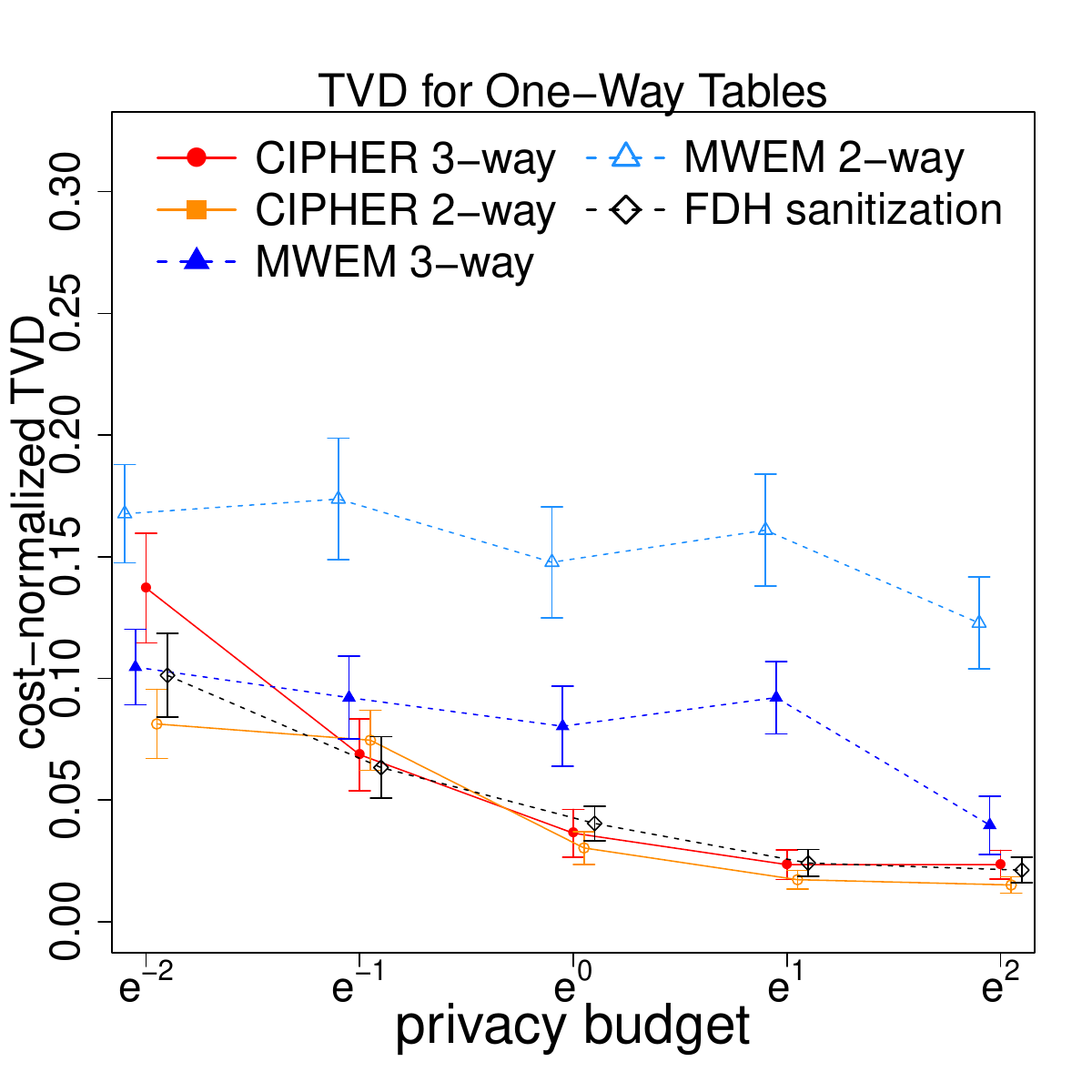}
\includegraphics[scale=0.205,trim={0.2cm 0.6cm 1cm 1.3cm},clip]{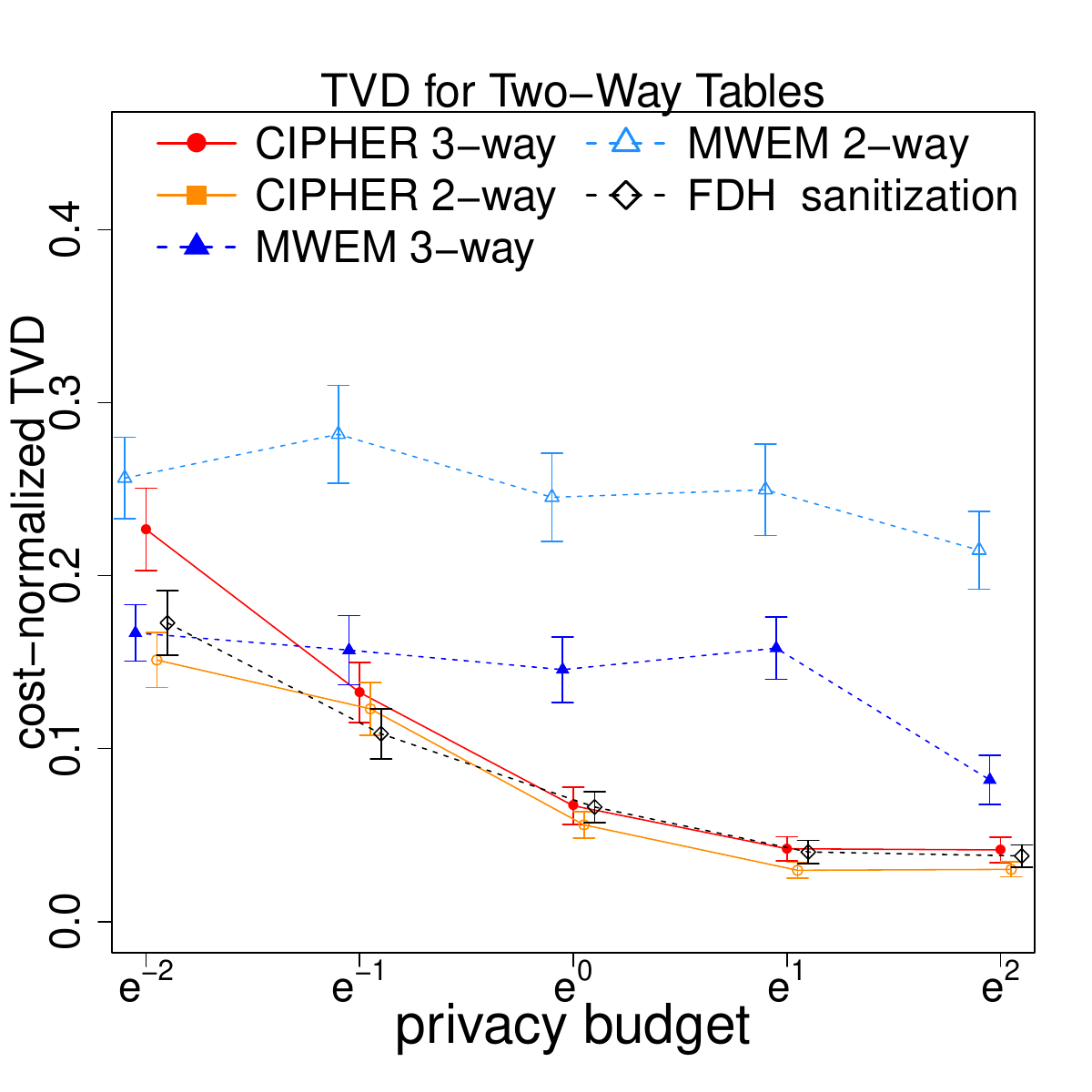}
\includegraphics[scale=0.205,trim={0.2cm 0.6cm 1cm 1.3cm},clip]{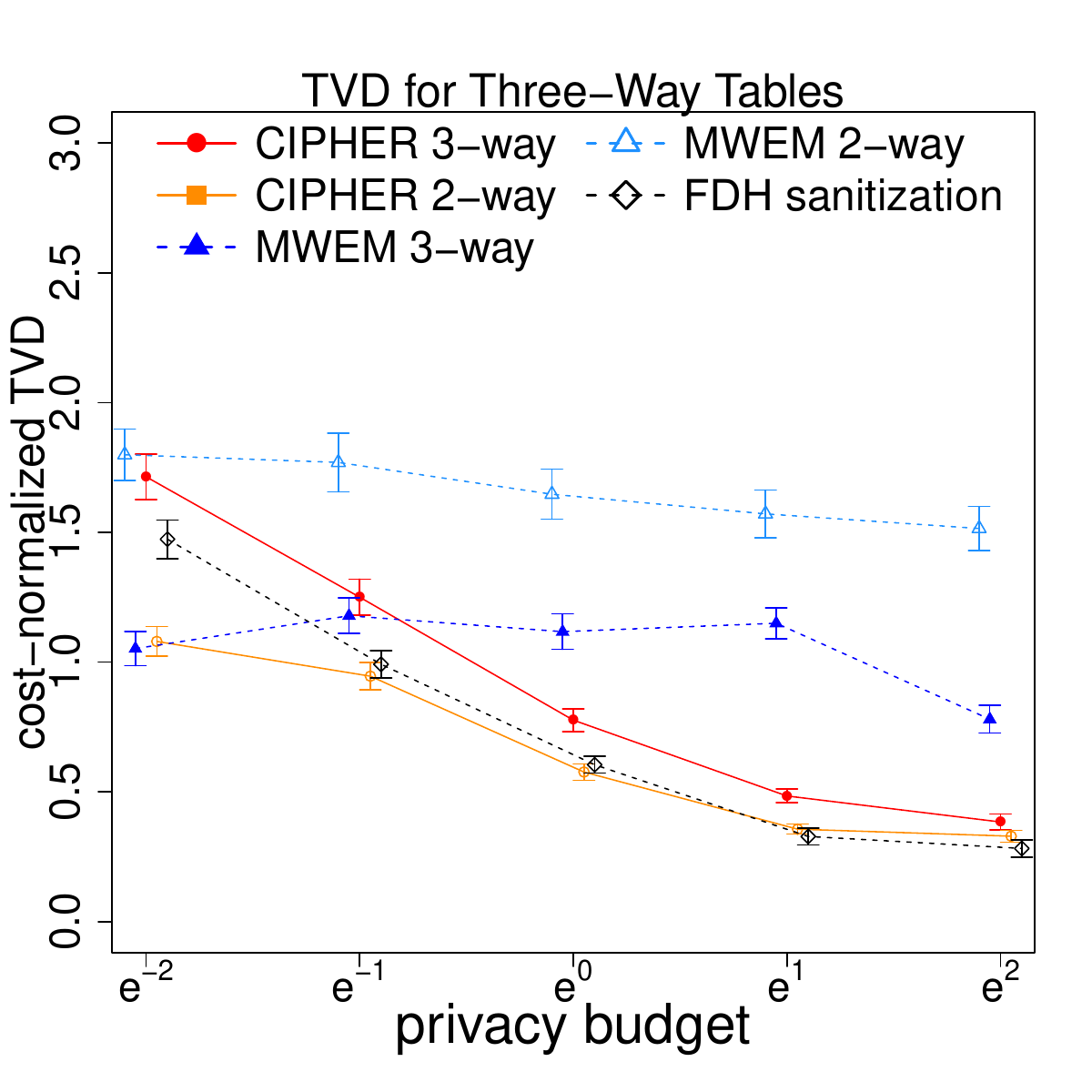}
\vspace{-9pt}
\caption{Cost-normalized total variation distance (mean $\pm$ SD over 1,000 repeats) (top: $n=200$; bottom: $n=500)$} \label{fig:simntvd} 
\end{figure}

In the first analysis, the average total variation distance (TVD) between the original and synthetic data sets was calculated for the 3-way, 2-way and 1-way marginals, respectively. Figure \ref{fig:simntvd} presents the results. After the cost normalization, CIPHER 2-way performs the best overall, especially for small $\epsilon$, and delivers similar performances as the FDH sanitization at large $\epsilon$. CIPHER 3-way is inferior to CIPHER 2-way. There is minimal change in the performance of MWEM produces across $\epsilon$.

In the second analysis, we examine the the $l_{\infty}$ error for $\mathcal{Q}_2$ and $\mathcal{Q}_3$, the results of which are given in Figure \ref{fig:simlinfty} for at $n=200$ (the findings are similar at $n=500$ and are available in the supplementary materials). CIPHER 2-way and the FDH sanitizations are similar with the former slightly better at $\epsilon=e^{-2}$. CIPHER 3-way is second-tier behind CIPHER 2-way and the FDH sanitization. The performance of MWEM does not seem to live up to the claim that it yields the optimal $l_{\infty}$ error for the set of queries that are fed to the algorithm \cite{hardt2012simple}. This could be due to the fact that $T$, which is not an easy hyperparameter to tune, was not optimized in a precise way (though roughly using independent data) in our implementation of MWEM.
\begin{figure}[!htb]
\vspace{-9pt}\centering
\includegraphics[scale=0.25,trim={0cm 0.7cm 0.5cm 1.3cm},clip]{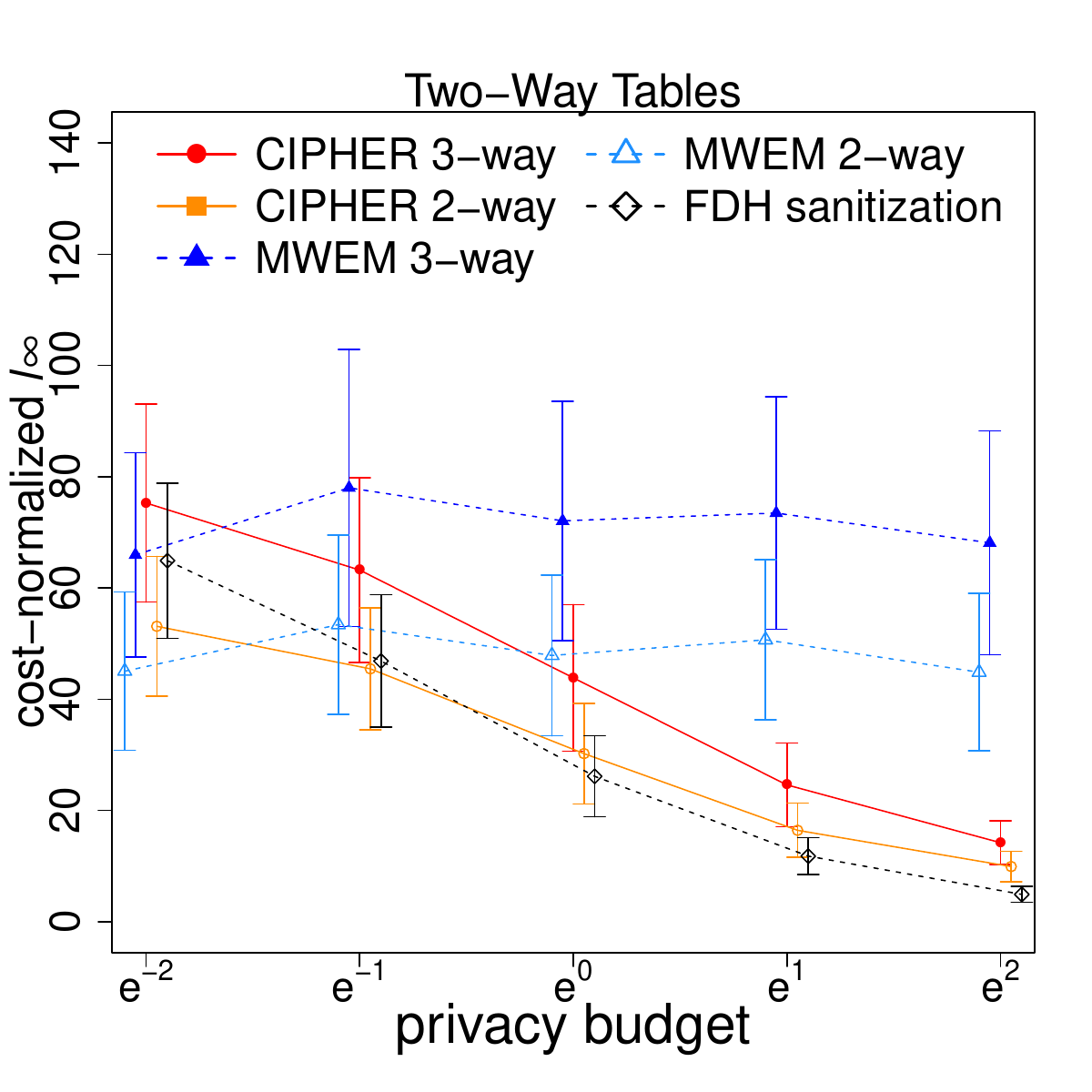}
\includegraphics[scale=0.25,trim={0cm 0.7cm 0.5cm 1.3cm},clip]{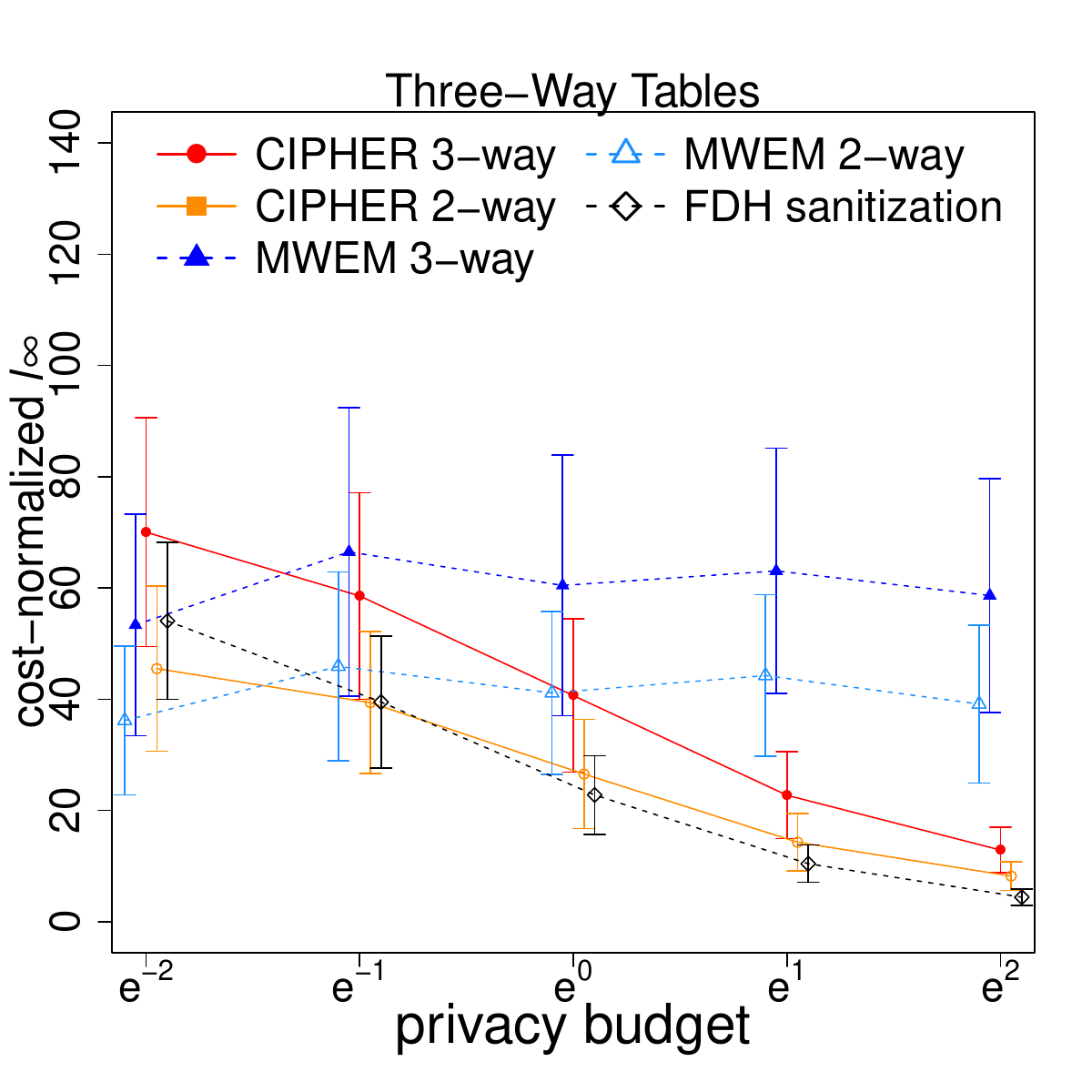}\\
\vspace{-9pt}\caption{Cost-normalized $l_{\infty}$ (mean $\pm$ SD over 1,000 repeats) at $n=200$} \label{fig:simlinfty}\vspace{-12pt}
\end{figure} 

In the third analysis, we fitted the multinomial logit model with a binary attribute as the outcome and the others as predictors. The inferences from the $m=5$ synthetic data sets were combined using the rule in \cite{liu2016model}. The bias, root mean square error (RMSE), coverage probability (CP) of the 95\% confidence interval (CI) were determined for each regression coefficient in the model. We present the results at $n=200$ and $\epsilon=e^{-2}, 1, e^2$ in Figure \ref{fig:simn200results} and those at $n=500$ and for $e^{-1}$ and $e$ at $n=200$ are listed in the supplementary materials. The observations are $n=500$ are consistent with $n=200$, and those at $\epsilon=e^{-1}$ and $e$ when $n=200$ are in between $e^{-2}$ and 1, and between $1$ and $e^2$, respectively in Figure \ref{fig:simn200results}. The 10 parameters from the model are listed on the $x$-axis. There is not much difference among CIPHER 2-way, CIPHER 3-way, and the FDH sanitization in cost-normalized bias or CP, but CIPHER 2-way delivers better performance in terms of cost-normalized RMSE (smaller) at all $\epsilon$ compared to CIPHER 3-way, and at small $\epsilon$ compared to the FDH sanitization. CIPHER and the FDH sanitization deliver near-nominal CP (95\%) across all the examined $\epsilon$ and both $n$ scenarios while MWEM suffers severe under-coverage on some parameters especially at small $\epsilon$. MWEM has the smallest cost-normalized RMSE for $\epsilon\le1$; but the RMSE values for CIPHER and the FDH sanitization catch up quickly and approach the original values as $\epsilon$ increases.
\begin{figure}[!hbt]
\centering
\includegraphics[scale=0.2,trim={0.1 2cm 1cm 0.2cm},clip]{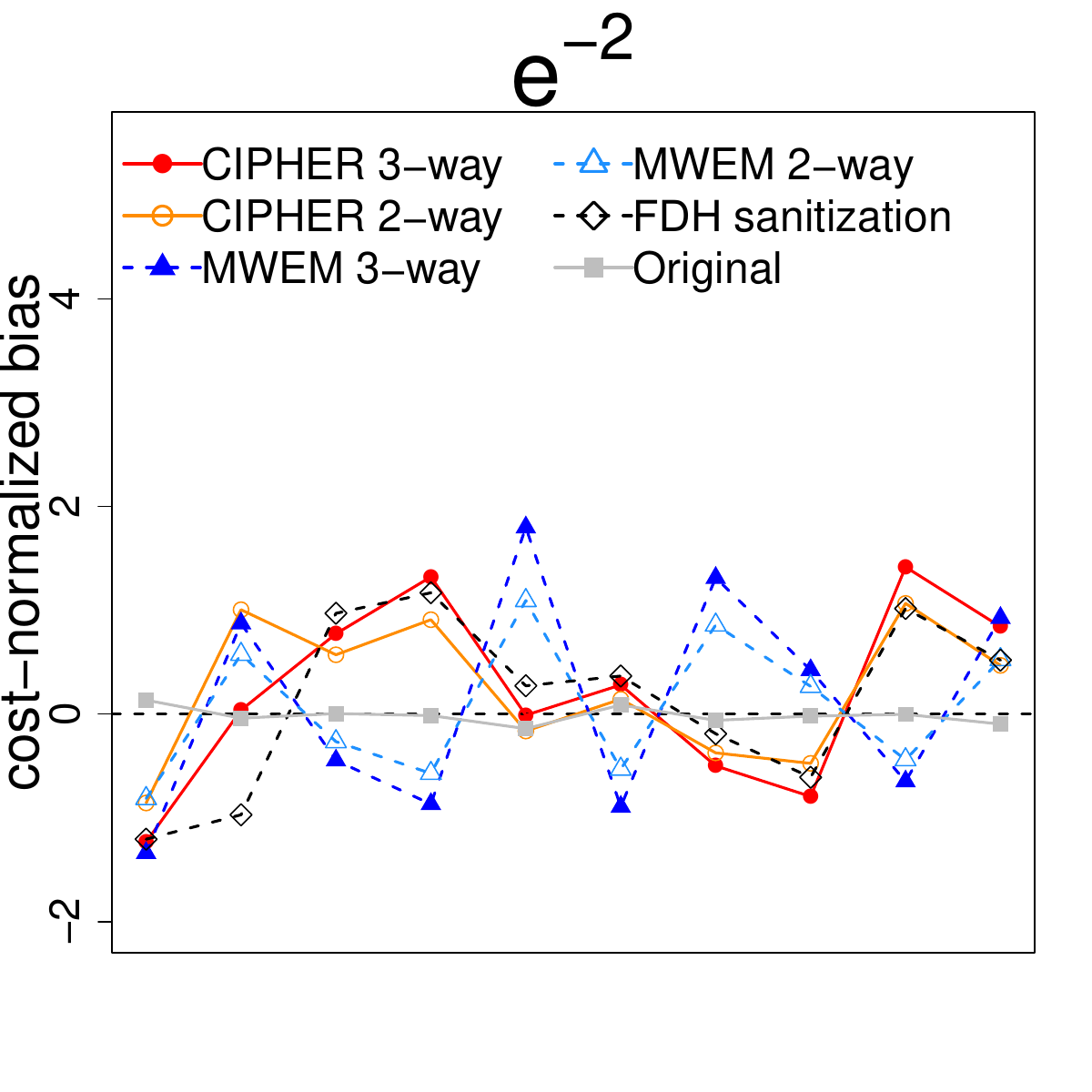}
\includegraphics[scale=0.2,trim={1.8cm 2cm 1cm 0.2cm},clip]{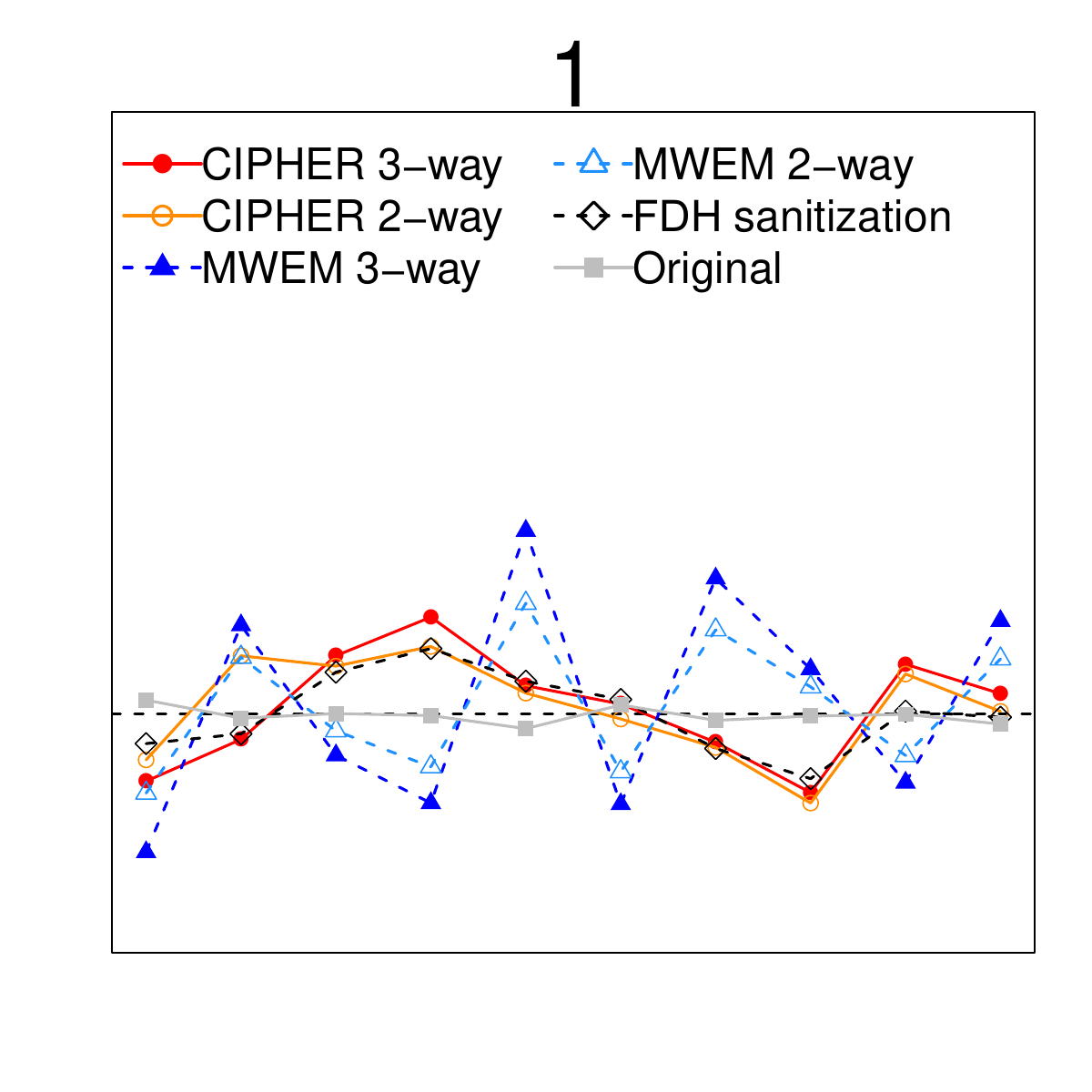}
\includegraphics[scale=0.2,trim={1.8cm 2cm 1cm 0.2cm},clip]{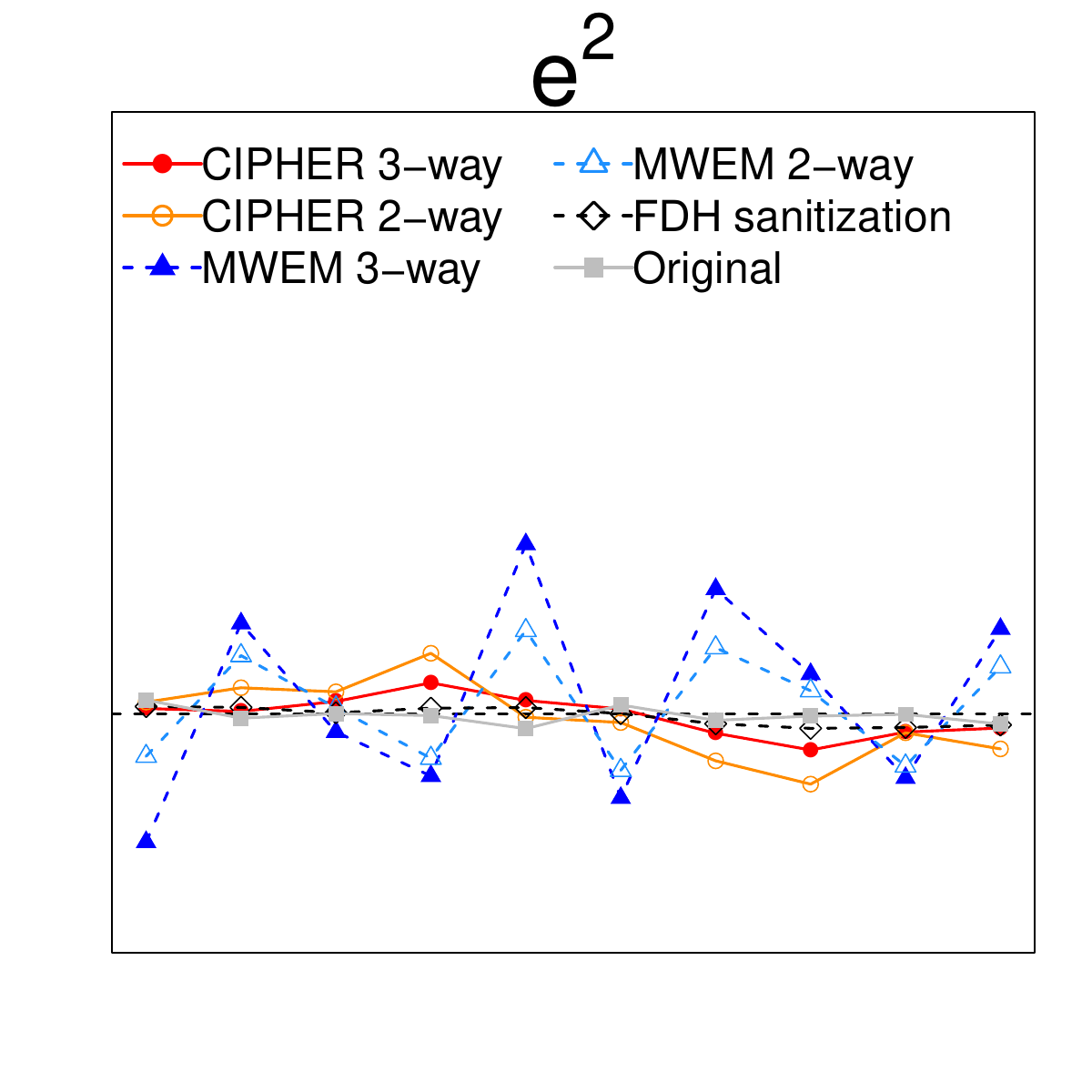}\\
\includegraphics[scale=0.2,trim={0.1 2cm 1cm 2cm},clip]{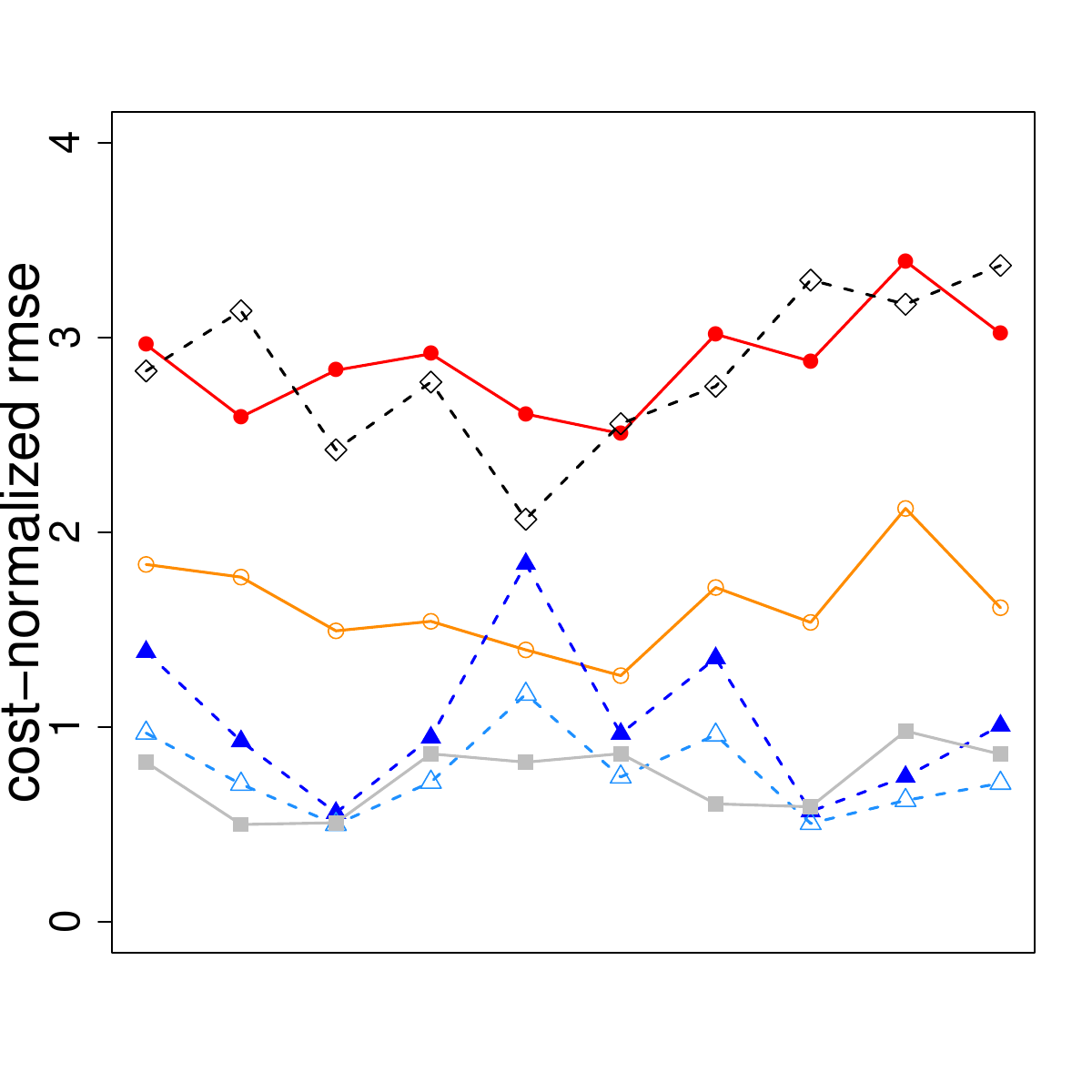} 
\includegraphics[scale=0.2,trim={1.8cm 2cm 1cm 2cm},clip]{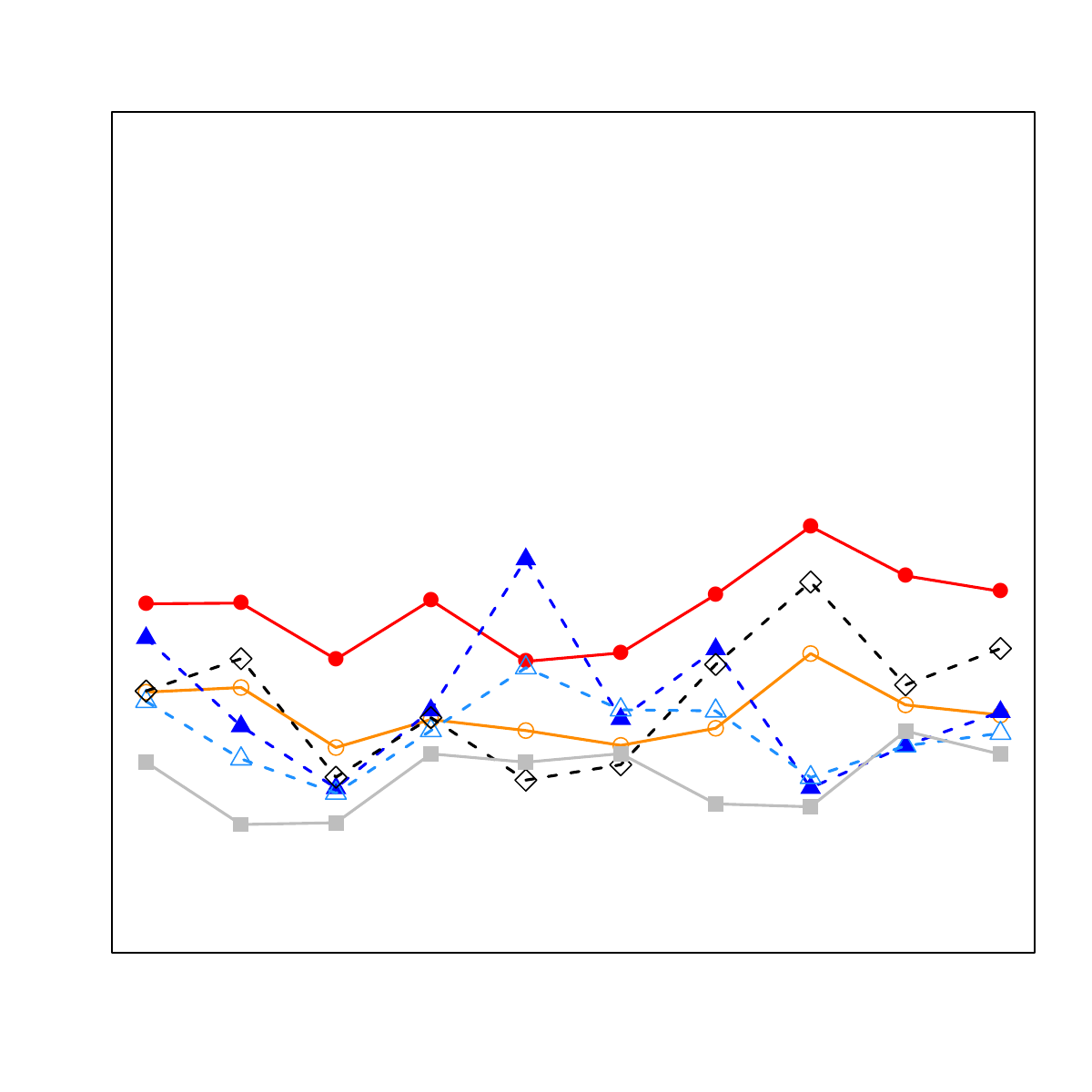}
\includegraphics[scale=0.2,trim={1.8cm 2cm 1cm 2cm},clip]{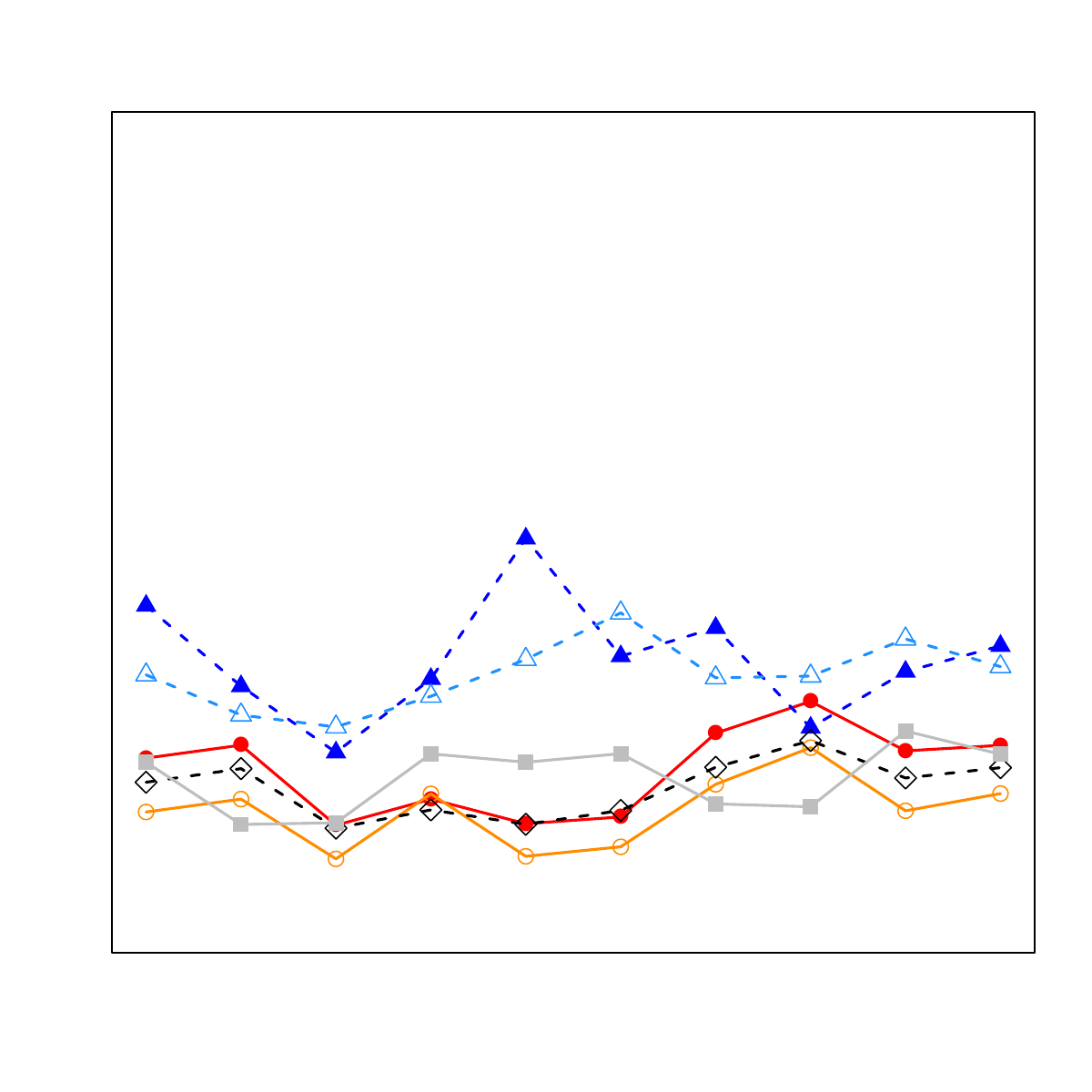}\\
\includegraphics[scale=0.2,trim={0.1 2cm 1cm 2cm},clip]{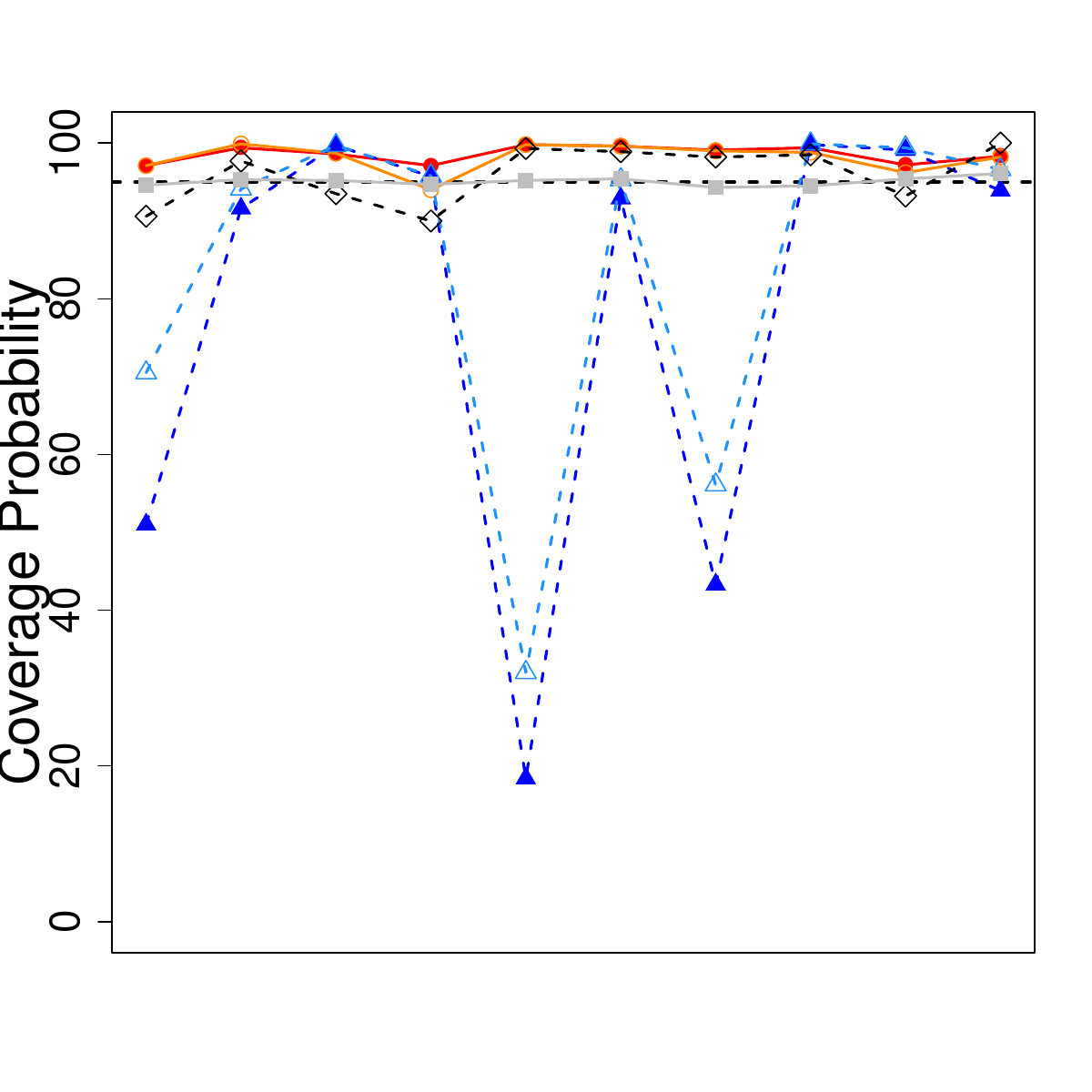} 
\includegraphics[scale=0.2,trim={1.8cm 2cm 1cm 2cm},clip]{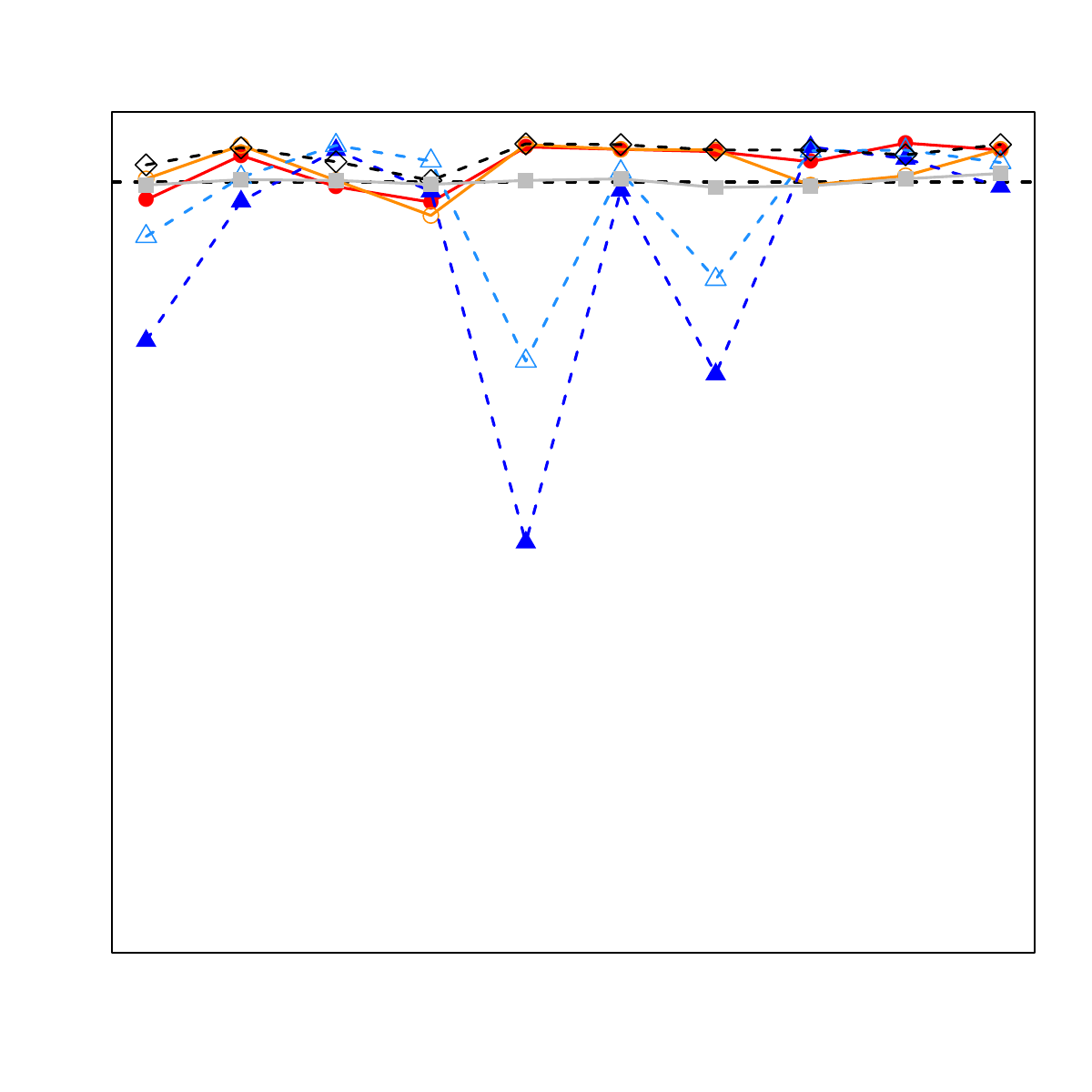} 
\includegraphics[scale=0.2,trim={1.8cm 2cm 1cm 2cm},clip]{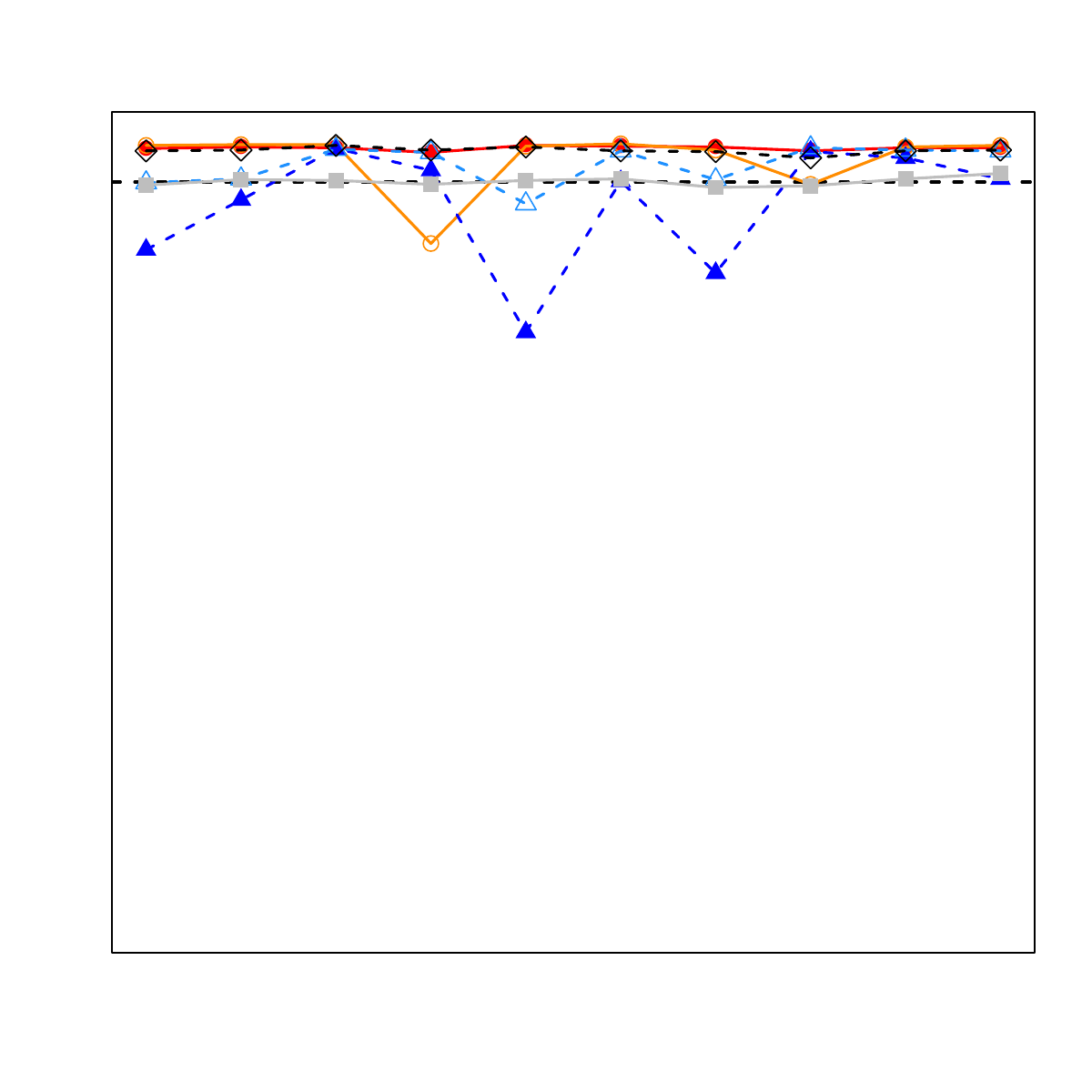} 
\vspace{-12pt}
\caption{Cost-normalized bias and root mean square error (rmse), and coverage probability (un-normalized) at $n=200$}\label{fig:simn200results}
\end{figure}

\begin{figure}[!htb]
\centering
$n=200$ \hspace{3.5cm} $n=500$\\
\includegraphics[scale=0.475,trim={0cm 0cm 0.5cm 1cm},clip]{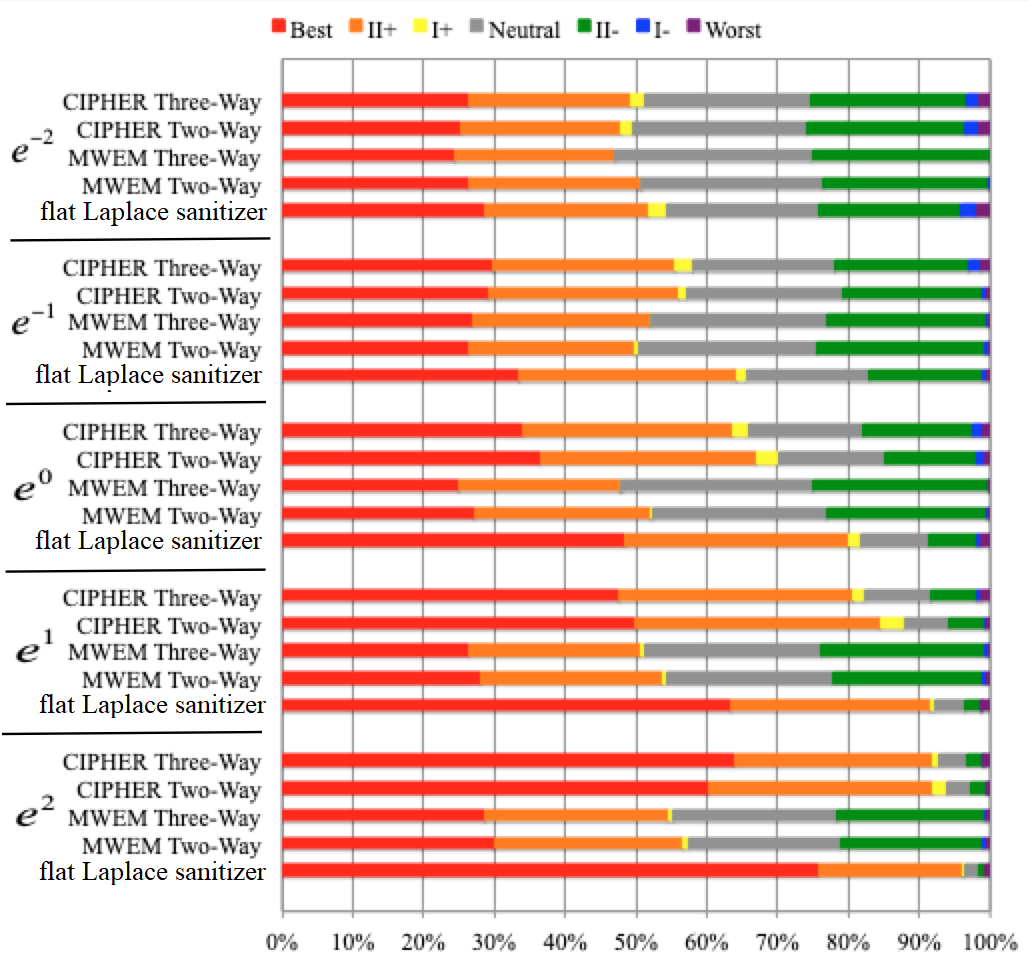}
\includegraphics[scale=0.475,trim={3.6cm 0cm 0.4cm 1cm},clip]{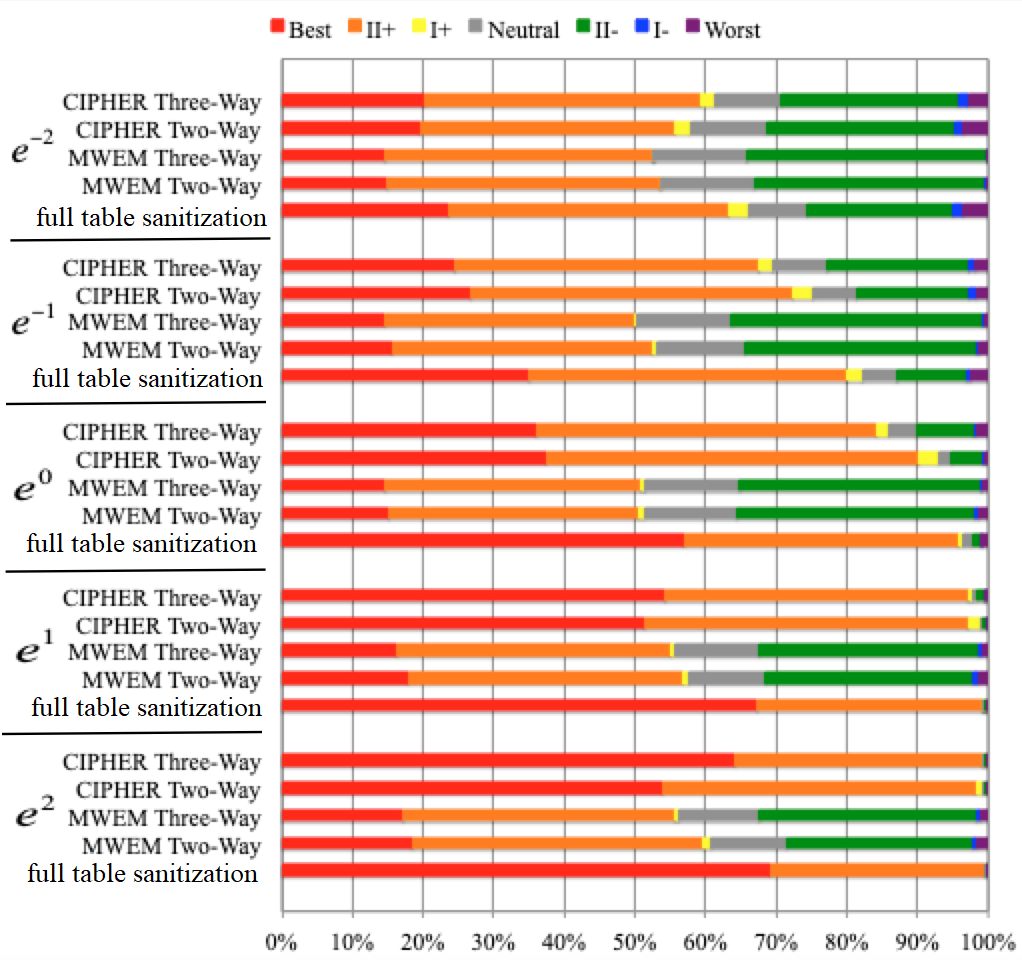}
\includegraphics[scale=0.7]{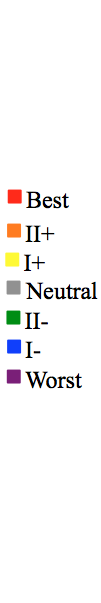} 
\vspace{-12pt}
\caption{The SSS (\textbf{S}igns and \textbf{S}tatistical \textbf{S}ignificance) assessment on the estimated regression coefficients for $n=200$ and $n=500$, un-normalized for storage cost}\label{fig:sssresults1}
\end{figure}

The results for the SSS assessment on the regression coefficients from the logistic regression are provided in Figure \ref{fig:sssresults1}, un-normalized for the storage cost. A method with the longest red bar (best-case scenario as defined in Table \ref{table:3S}) and the shortest purple bar (the worst-case scenario) would be preferable. The two inflated type II error (i.e., decreased power types) (the II+/orange bar and the I-/green bar) and neural (the gray bar) are acceptable, and the two inflated type I error types (I+/yellow bar and I-/blue bar) would preferably be of low probability. Per the listed criteria above, first, it is comforting to see the undesirable cases (purple+blue bars) are the shortest among all the 7 scenarios for each method. Second, the inferences improve quickly for CIPHER and the FDH sanitization and rather slowly for MWEM as $\epsilon$ increases. Third, the FDH sanitization is the best performer in preserving SSS, especially for the medium valued $\epsilon$, followed closely by CIPHER. Finally, even for CIPHER and the full table sanitization, there are always non-ignorable proportions of II+ (and II- when $\epsilon$ is small) especially when $\epsilon$ is as large as $e^2$, suggesting the sanitization decreases the efficiency of the statistical inferences, which is the expected price paid for privacy protection.

\begin{figure}[!htb]
\centering
$n=200$ \hspace{3cm} $n=500$\\
\includegraphics[scale=0.25,trim={0cm 0.5cm 1cm 2cm},clip]{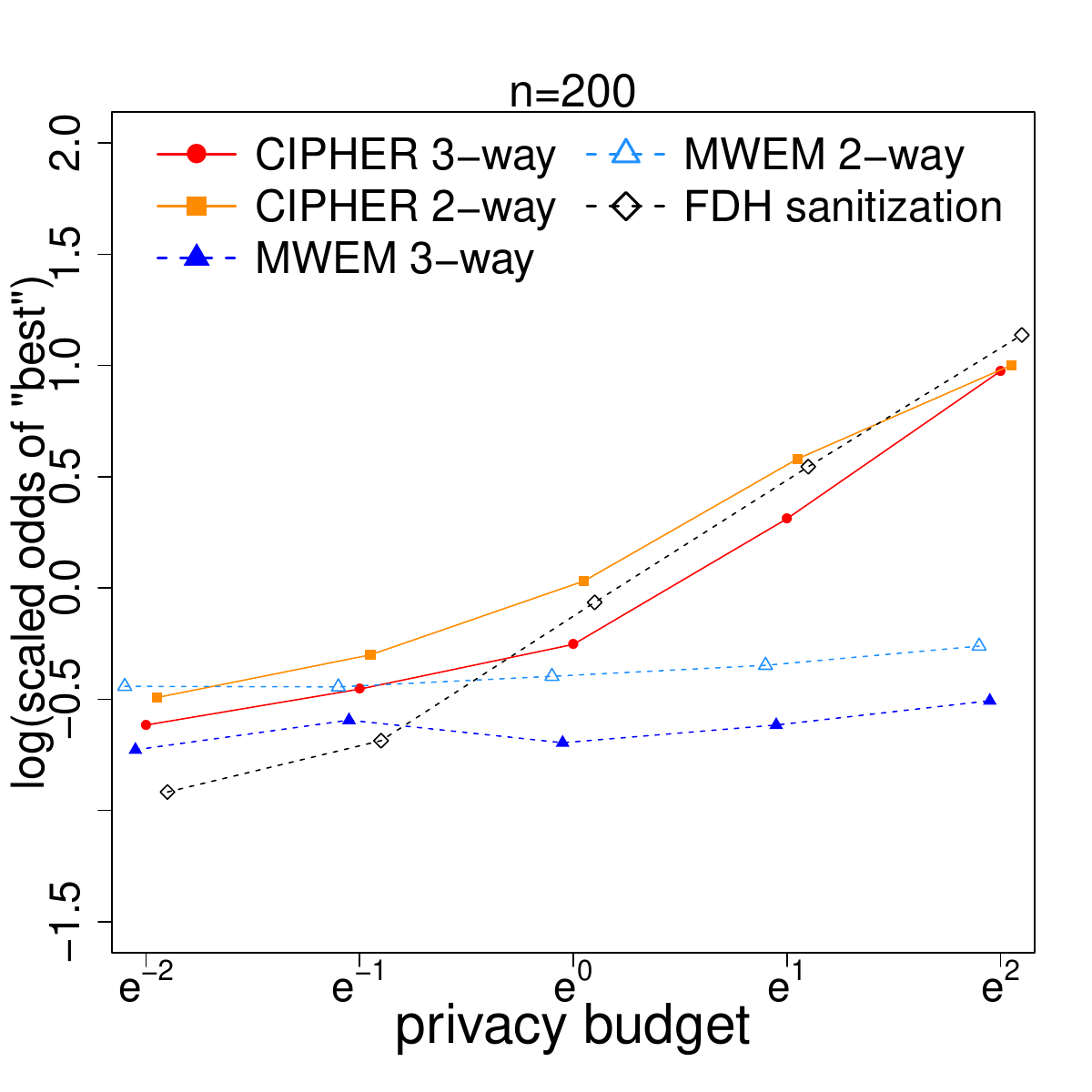} 
\includegraphics[scale=0.25,trim={0cm 0.5cm 1cm 2cm},clip]{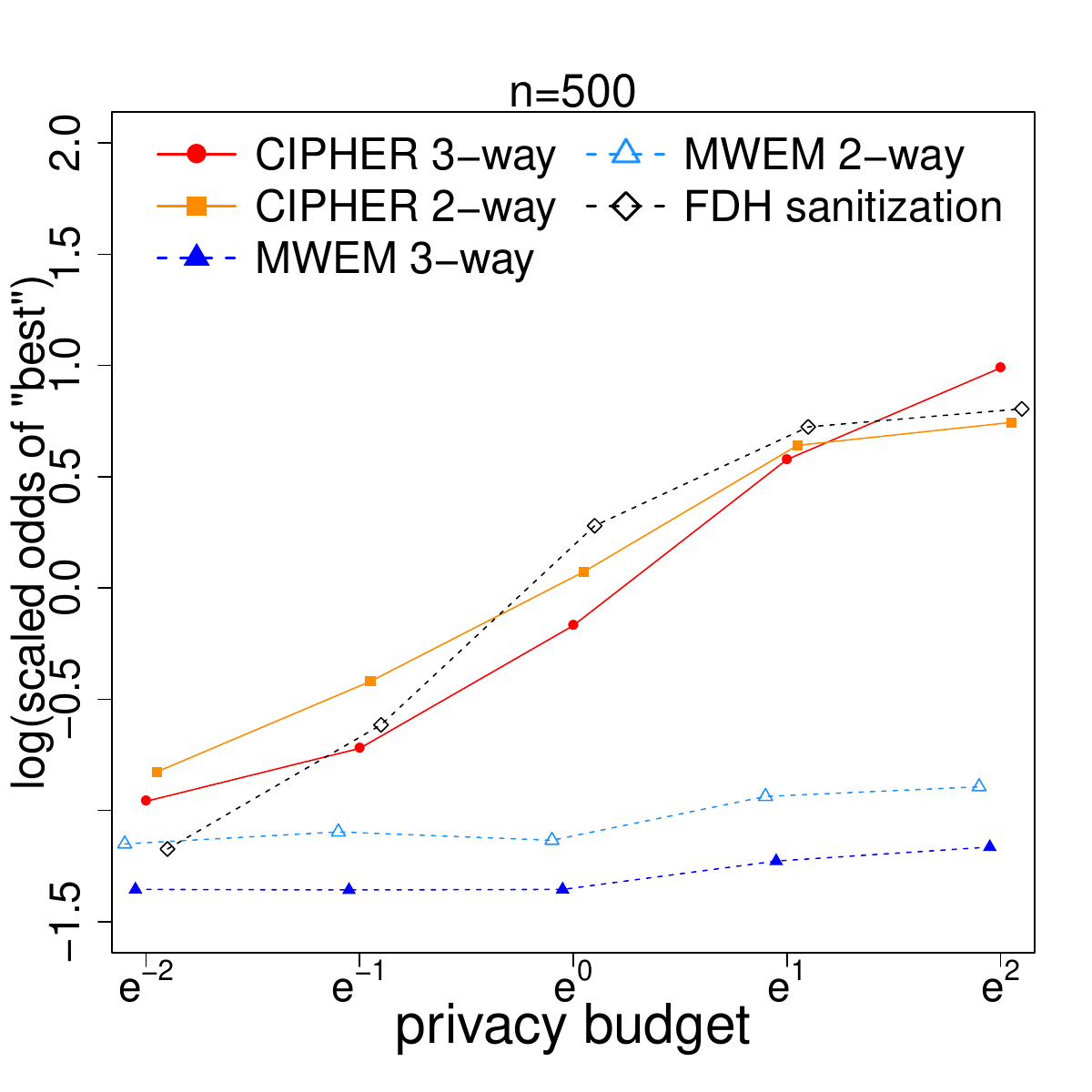}
\vspace{-12pt}
\caption{Cost-normalized log(odds of the ``best'' category) in the SSS assessment on the estimated regression coefficients}\label{fig:sssresults2}
\end{figure}

The cost-normalized log-odds of sanitized parameter estimates falling in the ``best'' category are provided in Figure \ref{fig:sssresults2} on the model parameters. The larger the odds, the more consistent the sanitized and the original inferences are on these parameters. Overall, the odds are similar for CIPHER 3-way, CIPHER 2-way, and the FDH sanitization, with CIPHER 3-way being the best at small $\epsilon$. MWEM performs similarly as the other methods at small $\epsilon$, but does not improve as $\epsilon$ increases.

\subsection{Experiment 2: Qualitative Bankruptcy Data} \label{sec:casestudy}\vspace{-3pt}
The experiments runs on a real-life qualitative bankruptcy data set. The data were collected to help identify the qualitative risk factors associated with bankru\-ptcy and is downloadable from the UCI Machine Learning repository \cite{Dua:2017}. The data set contains $n=250$ businesses and 7 variables. Though the data set does not contain any identifiers, sensitive information (such as credibility or bankruptcy status) can still be disclosed using the pseudo-identifiers left in the data (such as industrial risk level or competitiveness level), or be used to be linked to other public data to trigger other types of information disclosure. The supplementary materials provides a listing of the attributes in the data.

$\mathcal{Q}$ employed by the CIPHER and MWEM procedures contains one 4-way marginal, six 3-way marginals, and three 2-way marginals, that were selected based on the domain knowledge, and computational and analytical considerations when solving the linear equations in CIPHER, without referring to the actual values in the data. More details are provided in the supplementary materials on how $\mathcal{Q}$ was chosen.  The size of $\mathcal{Q}$ (the number of cell counts) is 149, which is about 10\% of the number of cells counts (1,458 cells) in the FDH sanitization. 

\begin{figure}[!htb]\centering
\includegraphics[scale=0.35]{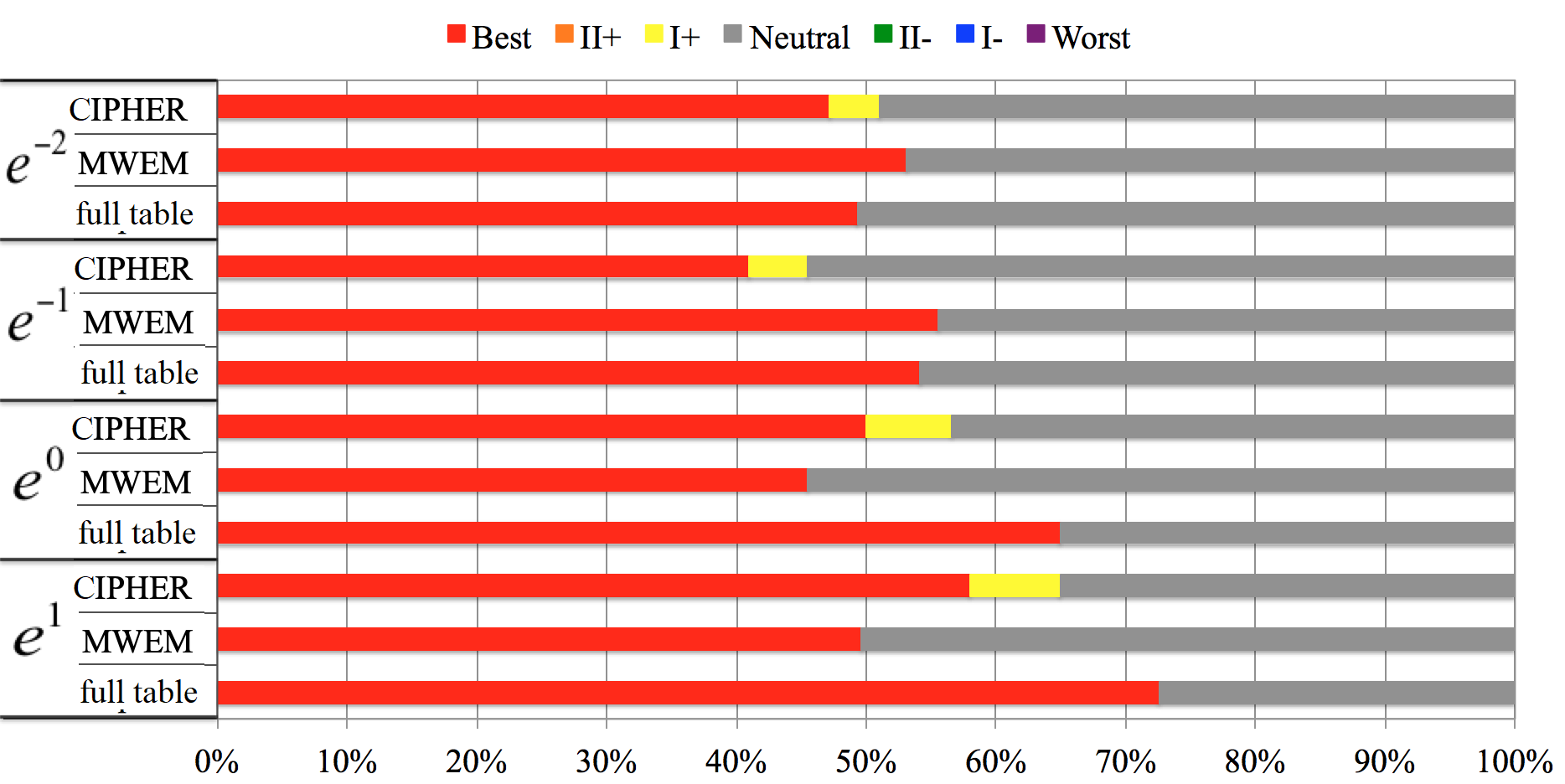} \vspace{-9pt}
\caption{The SSS assessment on the logistic regression coefficients in Experiment 2}\label{figure:caseresults}
\end{figure}

On the synthetic data generated by the three procedures, we ran a logistic regression model with ``Class'' as the outcome variable (bankruptcy vs non-bankruptcy) and examined its relationship with the other 6 qualitative categorical predictors \citep{kim2003discovery}. We applied the SSS assessment to the estimated parameters from the logistic regression and the results are presented in Figure \ref{figure:caseresults}. 
The figure suggests that all three methods perform well in the sense that the probability that they produced a ``bad'' estimate (the worst, II-, and I- categories) is close to 0, and the estimates are mostly likely to land in the ``best'' or the ``neutral'' categories. The FDH sanitization has the largest chance to produce estimates in the ``best'' category for $\epsilon \ge e^{-1}$, at a much higher storage cost ($\sim 8$ folds higher) than CIPHER and MWEM. MWEM has slightly better chance (50\%) landing in the ``best'' category when $\epsilon\le e^{-1}$ but does not improve for as $\epsilon$ increases. 

We also performed an SVM analysis to predict ``Class'' given the other attributes on a testing data (a random set of $50$ cases from the original). Per Table \ref{table:svm}, CIPHER is the obvious winner at $\epsilon\le 1$ with significantly better prediction accuracy than the other two and the accuracy is roughly constant.  FDH is better than CIPHER at $\epsilon>e$, but with a $\sim8$ -fold increase in the storage cost. The prediction accuracy remains $\sim50\%$ for MWEM across all examined $\epsilon$ values, basically not much better than a random guess on the outcome.
\begin{table}[!htb]
\caption{Prediction Accuracy (\%) on ``Class'' via SVM in Experiment 2}\label{table:svm}
\vspace{-9pt}
\centering
\resizebox{0.5\columnwidth}{!}{
\begin{tabular}{@{\hspace*{0.2cm}} l @{\hspace*{0.2cm}} | @{\hspace*{0.2cm}} c @{\hspace*{0.2cm}} c @{\hspace*{0.2cm}} c@{}}
\hline
$\epsilon$ & CIPHER & MWEM & FDH sanitization\\
\hline
$e^{-2}$ & \textbf{67.8} & 50.0 & 41.1 \\
$e^{-1}$ & \textbf{64.7} & 51.3 & 55.5 \\
1 & \textbf{68.5} & 51.0 & 63.8 \\
$e$ & 77.8 & 47.2 & \textbf{85.7} \\
$e^2$ & 90.3 & 47.3 & \textbf{98.8} \\
\hline
\end{tabular}}\vspace{-6pt}
\end{table}

\section{Discussion}\label{sec:discussion}\vspace{-2pt}
We propose the CIPHER procedure to generate differentially private empirical distributions from a set of low-order marginals. Once the empirical distributions are obtained, individual-level synthetic data can be generated. The experiment results implies that CIPHER delivers similar or superior performances to the FDH sanitization, especially at low privacy cost  after taking into account the storage cost. CIPHER in general delivers significantly better results on all the examined metrics than MWEM.  Both the CIPHER and MWEM procedures have multiple sources of errors in addition the sanitation randomness injected to ensure DP. For CIPHER, it is the shrinkage bias brought by the $l_2$ regularization; and for MWEM , it is the numerical errors introduced through the iterative procedure with a hard-to-choose $T$.  The asymptotic version of both CIPHER and MWEM is the FDH sanitization when the low-order marginals set contains only one query -- the full-dimensional table.  

We demonstrated the implementation of CIPHER for categorical data. The procedure also applies to data with numerical attributes, where the input would  be a set of low-dimensional histograms. This implies the numerical attributes will need to be cut into bins first before the application of CIPHER. After the sanitized empirical joint distribution is generated, values of the numerical attributes can be uniformly sampled from the sanitized bins. 

For future work, we plan to investigate the theoretical accuracy for CIPHER using some common utility measures (e.g., $l_\infty$ or $l_1$ errors); to apply CIPHER to data of higher dimensions in terms of both $p$ and the number of levels per attribute to examine the scalability of CIPHER; and to compare CIPHER with more methods that may also generate differentially private empirical distributions from a set of low-dimensional statistics, such as PrivBayes and the Fourier transform based method,  in both data utility and computational costs.

\section*{\large{Supplementary Materials}} \vspace{-3pt}
The supplementary materials are posted at \url{https://arxiv.org/abs/1812.05671}. The materials contains additional results from experiment 1, more details on the data used in experiment 2 and how $\mathcal{Q}$ is chosen, and the mathematical derivation of the linear equations sets $\boldsymbol{Ax}=\boldsymbol{b}$ for the three-variable and four-variable cases.

\vspace{-3pt}
\section*{\large{Acknowledgments}} \vspace{-3pt}
The authors would like to thank two anonymous reviewers for their comments and suggestions that helped improve the quality of the manuscript.

\vspace{12pt}
\bibliographystyle{spbasic_unsrt}
\bibliography{myref}

\end{document}


\mainmatter   
\title{Supplementary Materials to \\ \emph{Construction of Differentially Private Empirical Distributions from a  low-order Marginals Set through Solving Linear Equations with $l_2$ Regularization}}

\titlerunning{CIPHER} 
\author{Evercita Eugenio\inst{1} \and Fang Liu\inst{2}}
\tocauthor{Fang Liu and Evercita Eugenio}
\institute{Sandia National Laboratories, Livermore, CA 94550\\
\email{eceugen@sandia.gov}
\and
University of Notre Dame, Notre Dame, IN 46556, USA\\
\email{fang.liu.131@nd.edu}}

\footnotetext[1]{The research was funded by the US National Science Foundation grants \#1546373 and \#1717417.}

\maketitle
The supplementary materials contain additional simulation results in Experiment 1, more details on the data used in Experiment 2 and how $\mathcal{Q}$ is chosen, and the derivation of the linear equations sets $\boldsymbol{Ax}=\boldsymbol{b}$ for the three-variable and four-variable cases. Specifically,  
\begin{itemize}
\item  Fig. \ref{fig:simn200results} in Section 1 shows the results at all examined $\epsilon$ values for $n=200$, and  Fig. \ref{fig:simn500results} presents the results at $n=500$ in Experiment 1. 
\item Section 2 presents more details on the data used in Experiment 2 and how $\mathcal{Q}$ is chosen for CIPHER and MWEM. 

\item Section \ref{sec:fourvarderivation} includes the detailed derivation for $\boldsymbol{Ax}=\boldsymbol{b}$ using several examples when $p=3$ and $p=4$, respectively. The four-variable case  $p=4$ is also what was used in the CIPHER algorithm for the simulation study.
\end{itemize}

\section{Additional Results in Experiment 1}
\begin{figure}[!htb]
\vspace{-9pt}\centering
\includegraphics[scale=0.25,trim={0cm 0.7cm 0.5cm 1.3cm},clip]{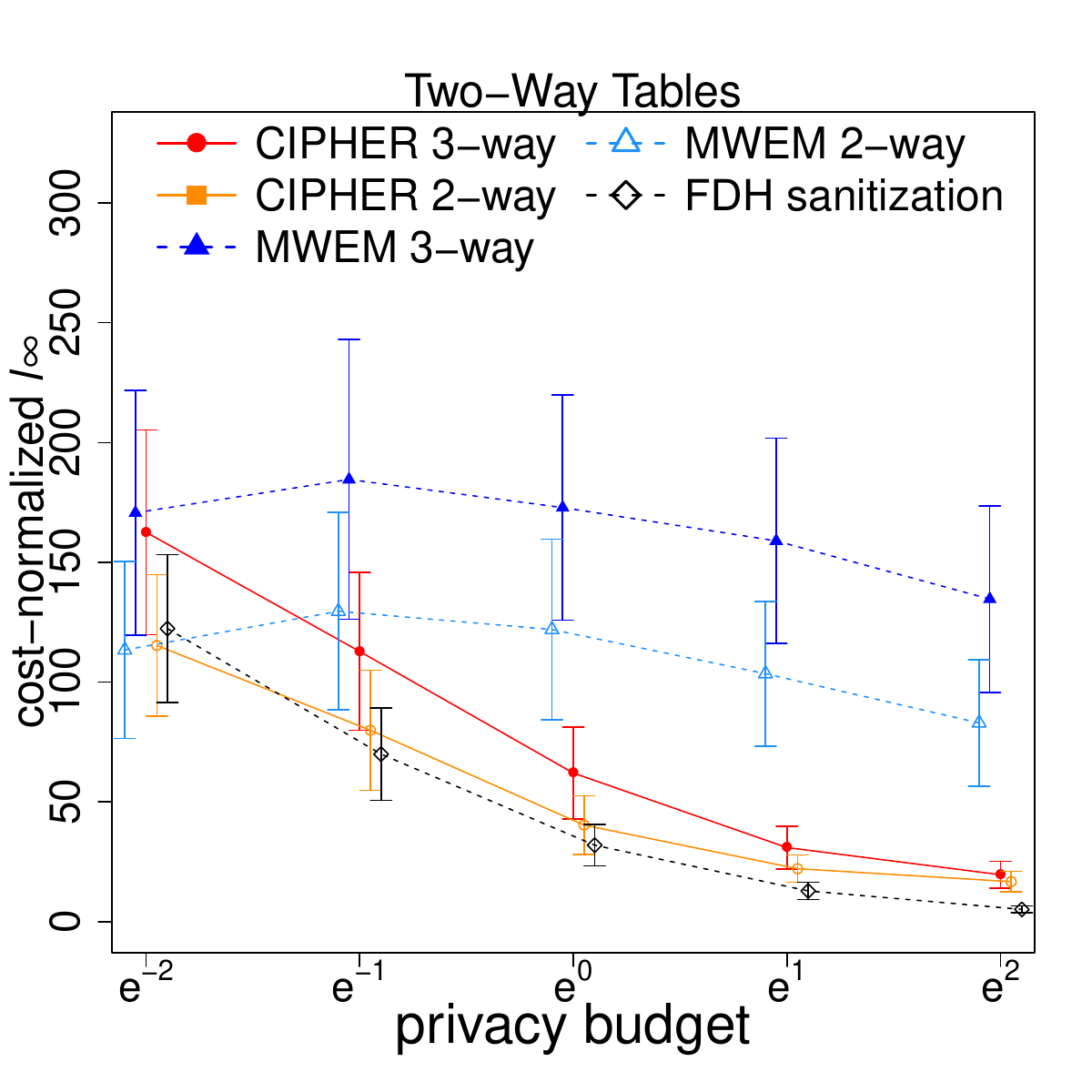}
\includegraphics[scale=0.25,trim={0cm 0.7cm 0.5cm 1.3cm},clip]{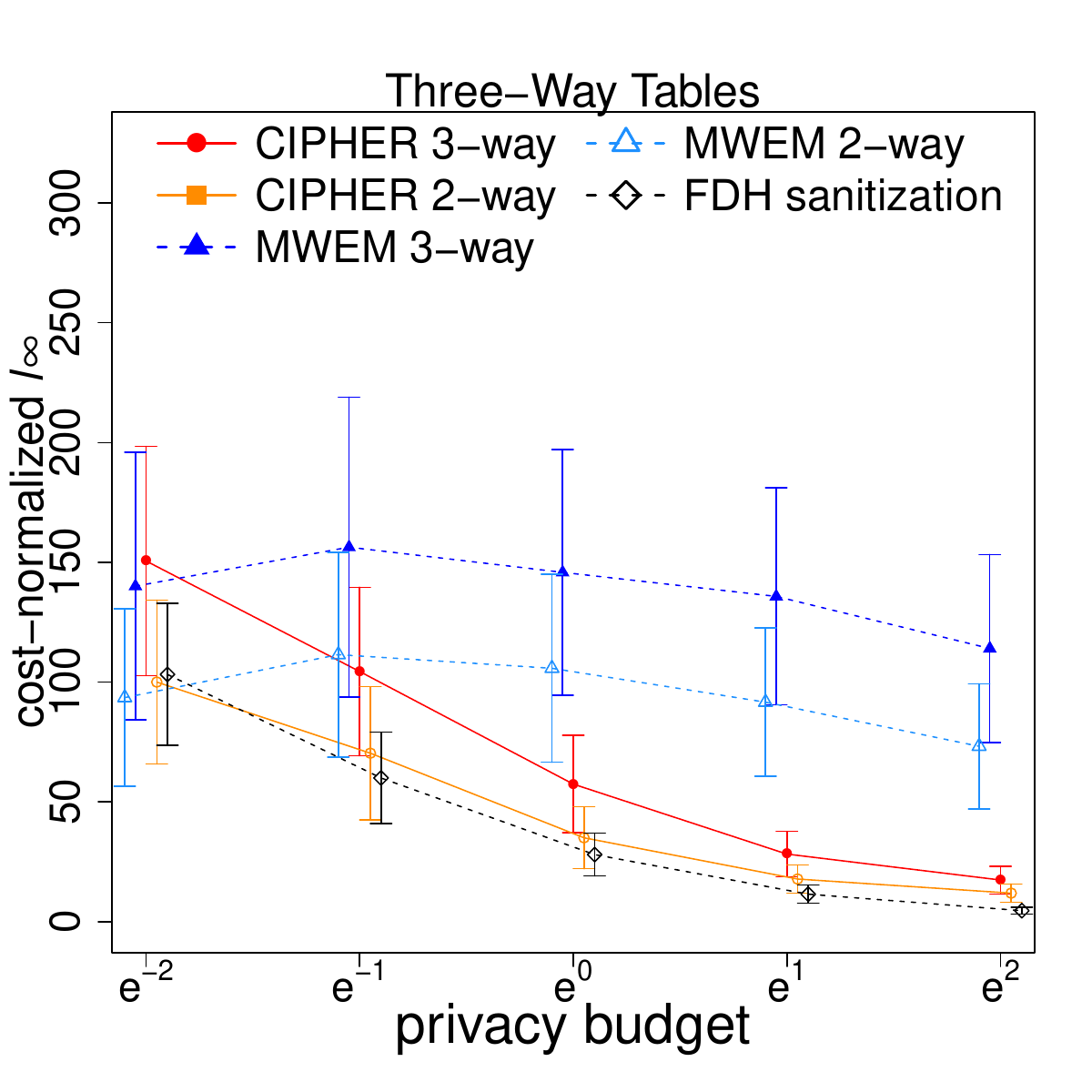}
\vspace{-9pt}\caption{Cost-normalized $l_{\infty}$ (mean $\pm$ SD over 1,000 repeats) at $n=500$} \label{fig:simlinfty}\vspace{-12pt}
\end{figure} 

\begin{landscape}
\begin{figure*}[!bt]
\centering
\includegraphics[scale=0.185,,trim={0.1 2cm 1.1cm 0.3cm},clip]{RevisedPlots/n200_bias_em2.pdf} 
\includegraphics[scale=0.185,,trim={1.8cm 2cm 1.1cm 0.3cm},clip]{RevisedPlots/n200_bias_em1.pdf} 
\includegraphics[scale=0.185,,trim={1.8cm 2cm 1.1cm 0.3cm},clip]{RevisedPlots/n200_bias_e0.pdf}
\includegraphics[scale=0.185,,trim={1.8cm 2cm 1.1cm 0.3cm},clip]{RevisedPlots/n200_bias_e1.pdf}
\includegraphics[scale=0.185,,trim={1.8cm 2cm 1.1cm 0.3cm},clip]{RevisedPlots/n200_bias_e2.pdf}\\
\includegraphics[scale=0.185,,trim={0.1 2cm 1.1cm 2cm},clip]{RevisedPlots/n200_cp_em2.pdf} 
\includegraphics[scale=0.185,,trim={1.8cm 2cm 1.1cm 2cm},clip]{RevisedPlots/n200_cp_em1.pdf} 
\includegraphics[scale=0.185,,trim={1.8cm 2cm 1.1cm 2cm},clip]{RevisedPlots/n200_cp_e0.pdf} 
\includegraphics[scale=0.185,,trim={1.8cm 2cm 1.1cm 2cm},clip]{RevisedPlots/n200_cp_e1.pdf} 
\includegraphics[scale=0.185,,trim={1.8cm 2cm 1.1cm 2cm},clip]{RevisedPlots/n200_cp_e2.pdf} \\
\includegraphics[scale=0.185,,trim={0.1 2cm 1.1cm 2cm},clip]{RevisedPlots/n200_rmse_em2.pdf} 
\includegraphics[scale=0.185,,trim={1.8cm 2cm 1.1cm 2cm},clip]{RevisedPlots/n200_rmse_em1.pdf} 
\includegraphics[scale=0.185,,trim={1.8cm 2cm 1.1cm 2cm},clip]{RevisedPlots/n200_rmse_e0.pdf}
\includegraphics[scale=0.185,,trim={1.8cm 2cm 1.1cm 2cm},clip]{RevisedPlots/n200_rmse_e1.pdf} 
\includegraphics[scale=0.185,,trim={1.8cm 2cm 1.1cm 2cm},clip]{RevisedPlots/n200_rmse_e2.pdf} \\
\includegraphics[scale=0.185,,trim={0.1 0.1cm 1.1cm 2cm},clip]{RevisedPlots/n200_ciwidth_em2.pdf}
\includegraphics[scale=0.185,,trim={1.8cm 0.1cm 1.1cm 2cm},clip]{RevisedPlots/n200_ciwidth_em1.pdf} 
\includegraphics[scale=0.185,,trim={1.8cm 0.1cm 1.1cm 2cm},clip]{RevisedPlots/n200_ciwidth_e0.pdf} 
\includegraphics[scale=0.185,,trim={1.8cm 0.1cm 1.1cm 2cm},clip]{RevisedPlots/n200_ciwidth_e1.pdf} 
\includegraphics[scale=0.185,,trim={1.8cm 0.1cm 1.1cm 2cm},clip]{RevisedPlots/n200_ciwidth_e2.pdf}
\vspace{-12pt}
\caption{Cost-normalized bias, cost-normalized root mean square error (rmse), coverage probability and cost-normalized  width of 95\% CI  at $n=200$ in Experiment 1}\label{fig:simn200results}
\end{figure*}

\begin{figure*}[!t]
\centering
\includegraphics[scale=0.185,trim={0.1 2cm 1.1cm 0.3cm},clip]{RevisedPlots/n500_bias_em2.pdf}
\includegraphics[scale=0.185,trim={1.8cm 2cm 1.1cm 0.3cm},clip]{RevisedPlots/n500_bias_em1.pdf} 
\includegraphics[scale=0.185,trim={1.8cm 2cm 1.1cm 0.3cm},clip]{RevisedPlots/n500_bias_e0.pdf}
\includegraphics[scale=0.185,trim={1.8cm 2cm 1.1cm 0.3cm},clip]{RevisedPlots/n500_bias_e1.pdf}
\includegraphics[scale=0.185,trim={1.8cm 2cm 1.1cm 0.3cm},clip]{RevisedPlots/n500_bias_e2.pdf}\\
\includegraphics[scale=0.185,trim={0.1 2cm 1.1cm 2cm},clip]{RevisedPlots/n500_cp_em2.pdf} 
\includegraphics[scale=0.185,trim={1.8cm 2cm 1.1cm 2cm},clip]{RevisedPlots/n500_cp_em1.pdf} 
\includegraphics[scale=0.185,trim={1.8cm 2cm 1.1cm 2cm},clip]{RevisedPlots/n500_cp_e0.pdf} 
\includegraphics[scale=0.185,trim={1.8cm 2cm 1.1cm 2cm},clip]{RevisedPlots/n500_cp_e1.pdf} 
\includegraphics[scale=0.185,trim={1.8cm 2cm 1.1cm 2cm},clip]{RevisedPlots/n500_cp_e2.pdf} \\
\includegraphics[scale=0.185,trim={0.1 2cm 1.1cm 2cm},clip]{RevisedPlots/n500_rmse_em2.pdf} 
\includegraphics[scale=0.185,trim={1.8cm 2cm 1.1cm 2cm},clip]{RevisedPlots/n500_rmse_em1.pdf} 
\includegraphics[scale=0.185,trim={1.8cm 2cm 1.1cm 2cm},clip]{RevisedPlots/n500_rmse_e0.pdf}
\includegraphics[scale=0.185,trim={1.8cm 2cm 1.1cm 2cm},clip]{RevisedPlots/n500_rmse_e1.pdf} 
\includegraphics[scale=0.185,trim={1.8cm 2cm 1.1cm 2cm},clip]{RevisedPlots/n500_rmse_e2.pdf} \\
\includegraphics[scale=0.185,trim={0.1 0.1cm 1.1cm 2cm},clip]{RevisedPlots/n500_ciwidth_em2.pdf}
\includegraphics[scale=0.185,trim={1.8cm 0.1cm 1.1cm 2cm},clip]{RevisedPlots/n500_ciwidth_em1.pdf} 
\includegraphics[scale=0.185,trim={1.8cm 0.1cm 1.1cm 2cm},clip]{RevisedPlots/n500_ciwidth_e0.pdf} 
\includegraphics[scale=0.185,trim={1.8cm 0.1cm 1.1cm 2cm},clip]{RevisedPlots/n500_ciwidth_e1.pdf} 
\includegraphics[scale=0.185,trim={1.8cm 0.1cm 1.1cm 2cm},clip]{RevisedPlots/n500_ciwidth_e2.pdf}
\vspace{-12pt}
\caption{Cost-normalized bias, cost-normalized root mean square error (rmse), coverage probability and cost-normalized  width of 95\% CI  at $n=500$ in Experiment 1.}\label{fig:simn500results}
\end{figure*}

\end{landscape}

\clearpage

\section{Details on the Bankruptcy Data in Experiment 2} \label{sec:fourvarderivation} 
The data set in Experiment 2 contains $n=250$ businesses and 7 variables listed in Table \ref{table:qbvariables}. 
\begin{table}[!htb]
\centering
\caption{Variables in the Bankruptcy Data}\label{table:qbvariables}
\begin{tabular}{@{}l@{\hskip 6pt}l@{}}
\hline
Variable &  Category (Frequency) \\
\hline
industrial risk (IR) & positive (80),average (89),negative (81)) \\
management risk (MR) & positive (62),average (119),negative (69) \\
financial flexibility (FF) & positive (57),average (119),negative (74) \\
credibility (CR) & positive (79),average (94),negative (77)\\
competitiveness (CO) & positive (91),average (103),negative (56) \\
operating risk (OR) & positive (79),average (114),negative (57)\\
Class & bankruptcy (107),non-bankruptcy (143) \\
\hline
\end{tabular}
\end{table}

$\mathcal{Q}$ employed by the CIPHER and MWEM procedures contains one 4-way contingency table, six 3-way contingency tables, and three 2-way contingency tables, generated as follows. We first generated a 6-category  Class/CR variable from the full cross-tabulation, both of which can be regarded as sensitive information and are expected to be associated, and a 9-category IR/CO variable from the cross-tabulation of the two which is also expected to correlated; and then applied the CIPHER 2-way and MWEM 2-way to the 5 variables with 6 (Class/CR), 9 (IR/CO), 3 (OR), 3 (MR), and 3 (FF) levels respectively. The size of $\mathcal{Q}$ (the number of cell counts) is 149 over 10 sets of 2D histogram queries.  After the synthetic data were generated, we decoupled the two sets of combined variables (Class/CR and IR/CO),  so the final synthetic data set still contain all 7 attributes as in the original data set. 

\clearpage
\begin{landscape}
\section{Derivation of Linear Equation sets $\mathbf{Ax=b}$}\label{sec:fourvarderivation}
In this section, we illustrate the derivation of the linear equation set given a pre-specified query set $\mathcal{Q}$ in the following three scenarios: (1) 3-variable $2 \times 2 \times 2$ case with  $\mathcal{Q}$ = all 2-way histograms; 2) 3-variable $2 \times 3 \times 3$ case with  $\mathcal{Q}$ = all 2-way histograms; 3) 4 variable case: $2 \times 2 \times 3 \times 3$ with  $\mathcal{Q}$ = all 2-way histograms.

\subsection{Three variable case $2 \times 2 \times 2$}
In the 3 variable case $2 \times 2 \times 2$, we first obtain
\begin{align*}
P(V_3=0|V_1)&= P(V_3=0,V_2=0|V_1)+ P(V_3=0,V_2=1|V_1)\\
&= P(V_3=0|V_2=0,V_1)P(V_2=0|V_1)+ P(V_3=0|V_2=1,V_1)P(V_2=1|V_1)\\
P(V_3=0|V_2)&= P(V_3=0,V_1=0|V_2)+ P(V_3=0,V_1=1|V_2)\\
&= P(V_3=0|V_1=0,V_2)P(V_1=0|V_2)+ P(V_3=0|V_1=1,V_2)P(V_1=1|V_2).
\end{align*}
Examining each scenario of $V_1$ and $V_2$, the two equations above can be expanded into four equations. 
\begin{equation}\label{eqn:222}
\begin{cases}
P(V_3=0|V_1=0) = P(V_3=0|V_2=0,V_1=0)P(V_2=0|V_1=0)+ P(V_3=0|V_2=1,V_1=0)P(V_2=1|V_1=0)\\
P(V_3=0|V_1=1) = P(V_3=0|V_2=0,V_1=1)P(V_2=0|V_1=1)+ P(V_3=0|V_2=1,V_1=1)P(V_2=1|V_1=1)\\
P(V_3=0|V_2=0) = P(V_3=0|V_1=0,V_2=0)P(V_1=0|V_2=0)+ P(V_3=0|V_1=1,V_2=0)P(V_1=1|V_2=0)\\
P(V_3=0|V_2=1) = P(V_3=0|V_1=0,V_2=1)P(V_1=0|V_2=1)+ P(V_3=0|V_1=1,V_2=1)P(V_1=1|V_2=1)
\end{cases}
\end{equation}
Using the sanitized values from the 2-way tables, then the left hand sides, denoted by $\mathbf{b}$ of the four equations above are known:
$\mathbf{b}=\left(P(V_3=0|V_1=0), P(V_3=0|V_1=1), P(V_3=0|V_2=0), P(V_3=0|V_2=1)\right)$.  Additionally, on the right hand sides, the elements of $P(V_2=0|V_1=0)$, $P(V_2=0|V_1=1)$, $P(V_1=0|V_2=0)$, $P(V_1=0|V_2=1)$, $P(V_2=1|V_1=0)$, $P(V_2=1|V_1=1)$, $P(V_1=1|V_2=0)$, and $P(V_1=1|V_2=1)$ can be calculated from the sanitized 2-way tables.  Therefore,  Eqn (\ref{eqn:222}) can be written as $\boldsymbol{b}=\boldsymbol{Az}$, where $\mathcal{z}=\left(P(V_3=0|V_1=0,V_2=0), P(V_3=0|V_1=1,V_2=0), P(V_3=0|V_1=0,V_2=1), P(V_3=0|V_1=1,V_2=1)\right)$ $\mathbf{A}$ contains known coefficients associated with $\mathbf{z}$. Note that though there are four equations in Eqn (\ref{eqn:222}), they actually are linearly dependent,  Therefore, we apply the Tikhonov regularization to solve for the four unknowns in $\mathbf{z}$. Once we get $\mathbf{z}$ $P(V_3=1|V_1,V_2)=1-P(V_3=0|V_1,V_2)$, we can subsequently calculate the joint probability among $(V_1,V_2,V_3)$ as in $P(V_1,V_2,V_3)=P(V_3|V_1,V_2)P(V_1,V_2)$, from which we can sample the synthetic data.

\subsection{Three variable case $2 \times 3 \times 3$}
In the 3 variable case ($2 \times 3 \times 3$), the initial equations are
\begin{align*}
P(V_3=0 | V_1) &= P(V_3=0, V_2=0 | V_1) + P(V_3=0, V_2=1 | V_1) + P(V_3=0, V_2=2 | V_1)\\
& = P(V_3=0 | V_2=0, V_1) P(V_2=0 | V_1) + P(V_3=0 | V_2=1, V_1) P(V_2=1 | V_1) + P(V_3=0 | V_2=2, V_1) P(V_2=2 | V_1) \\
P(V_3=0|V_2) &= P(V_3=0, V_1=0 | V_2) + P(V_3=0, V_1=1 | V_2) \\
&= P(V_3=0 | V_1=0, V_2) P(V_1=0 | V_2) + P(V_3=0 | V_1=1, V_2) P(V_1=1 | V_2)\\
P(V_3=1 | V_1) &= P(V_3=1, V_2=0 | V_1) + P(V_3=1, V_2=1 | V_1) + P(V_3=1, V_2=2 | V_1)\\
& = P(V_3=1 | V_2=0, V_1) P(V_2=0 | V_1) + P(V_3=1 | V_2=1, V_1) P(V_2=1 | V_1) + P(V_3=1 | V_2=2, V_1) P(V_2=2 | V_1) \\
P(V_3=1|V_2) &= P(V_3=1, V_1=0 | V_2) + P(V_3=1, V_1=1 | V_2) \\
&= P(V_3=1 | V_1=0, V_2) P(V_1=0 | V_2) + P(V_3=1 | V_1=1, V_2) P(V_1=1 | V_2)
\end{align*}
Examining each scenario of $V_1$ and $V_2$, we can expand the above 4 equations into 10 equations with 12 unknowns. The left sides of the 10 equations compose $\mathbf{b}$, and $\mathbf{z}$ comprises of the 12 unknowns, which are the conditional probabilities of $V_3=0|V_1,V_2$ and $V_3=1|V_1,V_2$, and $\mathbf{A}$ contains the corresponding coefficients. We apply the Tikhonov regularization to solve for $\mathbf{z}$ from $\mathbf{b=Az}$. 

\begin{align*}
P(V_3=0 | V_1=0) &= P(V_3=0 | V_2=0, V_1=0) P(V_2=0 | V_1=0) \\
& \qquad + P(V_3=0 | V_2=1, V_1=0) P(V_2=1 | V_1=0) + P(V_3=0 | V_2=2, V_1=0) P(V_2=2 | V_1=0) \numberthis \\
P(V_3=0 | V_1=1) &= P(V_3=0 | V_2=0, V_1=1) P(V_2=0 | V_1=1) \\
& \qquad + P(V_3=0 | V_2=1, V_1=1) P(V_2=1 | V_1=1) + P(V_3=0 | V_2=2, V_1=1) P(V_2=2 | V_1=1) \numberthis \\
P(V_3=0 | V_2=0) &= P(V_3=0 | V_1=0, V_2=0) P(V_1=0 | V_2=0) + P(V_3=0 | V_1=1, V_2=0) P(V_1=1 | V_2=0) \numberthis \\
P(V_3=0 | V_2=1) &= P(V_3=0 | V_1=0, V_2=1) P(V_1=0 | V_2=1) + P(V_3=0 | V_1=1, V_2=1) P(V_1=1 | V_2=1) \numberthis \\
P(V_3=0 | V_2=2) &= P(V_3=0 | V_1=0, V_2=2) P(V_1=0 | V_2=2) + P(V_3=0 | V_1=1, V_2=2) P(V_1=1 | V_2=2) \numberthis \\
P(V_3=1 | V_1=0) &= P(V_3=1 | V_2=0, V_1=0) P(V_2=0 | V_1=0) \\
& \qquad + P(V_3=1 | V_2=1, V_1=0) P(V_2=1 | V_1=0) + P(V_3=1 | V_2=2, V_1=0) P(V_2=2 | V_1=0) \numberthis \\
P(V_3=1 | V_1=1) &= P(V_3=1 | V_2=0, V_1=1) P(V_2=0 | V_1=1) \\
& \qquad + P(V_3=1 | V_2=1, V_1=1) P(V_2=1 | V_1=1) + P(V_3=1 | V_2=2, V_1=1) P(V_2=2 | V_1=1) \numberthis \\
P(V_3=1 | V_2=0) &= P(V_3=1 | V_1=0, V_2=0) P(V_1=0 | V_2=0) + P(V_3=1 | V_1=1, V_2=0) P(V_1=1 | V_2=0) \numberthis \\
P(V_3=1 | V_2=1) &= P(V_3=1 | V_1=0, V_2=1) P(V_1=0 | V_2=1) + P(V_3=1 | V_1=1, V_2=1) P(V_1=1 | V_2=1) \numberthis \\
P(V_3=1 | V_2=2) &= P(V_3=1 | V_1=0, V_2=2) P(V_1=0 | V_2=2) + P(V_3=1 | V_1=1, V_2=2) P(V_1=1 | V_2=2) \numberthis 
\end{align*}

\subsection{Four variable case $2 \times 2 \times 3 \times 3$}
In this example, the variables $V_1$ and $V_2$ have two categories, and $V_3$ and $V_4$ have three categories.  We assume the query set $\mathcal{Q}$ consists of all 2D histograms among he variables of $V_1$, $V_2$, $V_3$ and $V_4$ (The procedures are similar if $\mathcal{Q}$ consists of other types of histograms, such as all 3D histograms, and a mixture of 2D or 3D histograms). CIPHER first solves for the probability distribution for all 3D histograms given 2D histograms, the procedures are similar to the 3-variable examples in Sections 2.1 and 2.2 of the supplementary materials. Once the 3D histograms are available, we can calculate the probability distribution of the four variable is calculated given the 3D histograms.

The initial equations are
\begin{align*}
P(V_4=0 | V_1, V_2) &= P(V_4=0, V_3=0 | V_1, V_2) + P(V_4=0, V_3=1 | V_1, V_2) + P(V_4=0, V_3=2 | V_1, V_2)\\
				 &= P(V_4=0 | V_3=0, V_1, V_2) P(V_3=0 | V_1, V_2) + P(V_4=0 | V_3=1, V_1, V_2) P(V_3=1 | V_1, V_2) \\
				 & \qquad + P(V_4=0 | V_3=2, V_1, V_2) P(V_3=2 | V_1, V_2) \\
P(V_4=0 |V_1, V_3) &= P(V_4=0, V_2=0 | V_1, V_3) + P(V_4=0, V_2=1 | V_1, V_3)\\
				 &= P(V_4=0 | V_2=0, V_1, V_3) P(V_2=0 | V_1, V_3) + P(V_4=0 | V_2=1, V_1, V_3) P(V_2=1 | V_1, V_3) \\
P(V_4=0 |V_2, V_3) &= P(V_4=0, V_1=0 | V_2, V_3) + P(V_4=0, V_1=1 | V_2, V_3)\\
				 &= P(V_4=0 | V_1=0, V_2, V_3) P(V_1=0 | V_2, V_3) + P(V_4=0 | V_1=1, V_2, V_3) P(V_1=1 | V_2, V_3) \\	 
P(V_4=1 | V_1, V_2) &= P(V_4=1, V_3=0 | V_1, V_2) + P(V_4=1, V_3=1 | V_1, V_2) + P(V_4=1, V_3=2 | V_1, V_2) \\
				 &= P(V_4=1 | V_3=0, V_1, V_2) P(V_3=0 | V_1, V_2) + P(V_4=1 | V_3=1, V_1, V_2) P(V_3=1 | V_1, V_2) \\
				 &\qquad + P(V_4=1 | V_3=2, V_1, V_2) P(V_3=2 | V_1, V_2) \\
P(V_4=1 |V_1, V_3) &= P(V_4=1, V_2=0 | V_1, V_3) + P(V_4=1, V_2=1 | V_1, V_3)\\
				 &= P(V_4=1 | V_2=0, V_1, V_3) P(V_2=0 | V_1, V_3) + P(V_4=1 | V_2=1, V_1, V_3) P(V_2=1 | V_1, V_3) \\
P(V_4=1 |V_2, V_3) &= P(V_4=1, V_1=0 | V_2, V_3) + P(V_4=1, V_1=1 | V_2, V_3)\\
				 &= P(V_4=1 | V_1=0, V_2, V_3) P(V_1=0 | V_2, V_3) + P(V_4=1 | V_1=1, V_2, V_3) P(V_1=1 | V_2, V_3) 
\end{align*}
Examining each scenario of $V_1$, $V_2$, and $V_3$, we can expand the above 6 equations into 32 with 24 unknowns (the conditional probability. Again, 32 equations are linearly dependent, and its rank is $<24$. Therefore, we apply the Tikhonov regularization to solve for $\mathbf{z}$ from $\mathbf{b=Az}$, where $\mathbf{b}$ contains the left sides of the 32 equations, and $\mathbf{z}$ refer to the conditional probabilities of  $V_4=0|V_1,V_2,V_3$ and $V_4=1|V_1,V_2,V_3$, and $\mathbf{A}$ contains the corresponding coefficients. 

\begin{align*}
P(V_4=0 | V_1=0, V_2=0) &= P(V_4=0 | V_1=0, V_2=0, V_3=0) P(V_3=0 | V_1=0, V_2=0) \\
	& \qquad + P(V_4=0 | V_1=0, V_2=0, V_3=1) P(V_3=1 | V_1=0, V_2=0) \numberthis \\
	& \qquad + P(V_4=0 | V_1=0, V_2=0, V_3=2) P(V_3=2 | V_1=0, V_2=0) \\
P(V_4=0 | V_1=1, V_2=0) &= P(V_4=0 | V_1=1, V_2=0, V_3=0) P(V_3=0 | V_1=1, V_2=0) \\
	& \qquad  + P(V_4=0 | V_1=1, V_2=0, V_3=1) P(V_3=1 | V_1=1, V_2=0) \numberthis \\
	& \qquad  + P(V_4=0 | V_1=1, V_2=0, V_3=2) P(V_3=2 | V_1=1, V_2=0) \\
P(V_4=0 | V_1=0, V_2=1) &= P(V_4=0 | V_1=0, V_2=1, V_3=0) P(V_3=0 | V_1=0, V_2=1) \\
	& \qquad  + P(V_4=0 | V_1=0, V_2=1, V_3=1) P(V_3=1 | V_1=0, V_2=1) \numberthis \\
	& \qquad  + P(V_4=0 | V_1=0, V_2=1, V_3=2) P(V_3=2 | V_1=0, V_2=1) \\
P(V_4=0 | V_1=1, V_2=1) &= P(V_4=0 | V_1=1, V_2=1, V_3=0) P(V_3=0 | V_1=1, V_2=1) \\
	& \qquad  + P(V_4=0 | V_1=1, V_2=1, V_3=1) P(V_3=1 | V_1=1, V_2=1) \numberthis \\
	& \qquad  + P(V_4=0 | V_1=1, V_2=1, V_3=2) P(V_3=2 | V_1=1, V_2=1)\\
P(V_4=0 | V_1=0, V_3=0) &= P(V_4=0 | V_1=0, V_2=0, V_3=0) P(V_2=0 | V_1=0, V_3=0) \\
	& \qquad  + P(V_4=0 | V_1=0, V_2=1, V_3=0) P(V_2=1 | V_1=0, V_3=0) \numberthis \\
P(V_4=0 | V_1=1, V_3=0) &= P(V_4=0 | V_1=1, V_2=0, V_3=0) P(V_2=0 | V_1=1, V_3=0) \\
	& \qquad  + P(V_4=0 | V_1=1, V_2=1, V_3=0) P(V_2=1 | V_1=1, V_3=0) \numberthis \\
P(V_4=0 | V_1=0, V_3=1) &= P(V_4=0 | V_1=0, V_2=0, V_3=1) P(V_2=0 | V_1=0, V_3=1) \\
	& \qquad  + P(V_4=0 | V_1=0, V_2=1, V_3=1) P(V_2=1 | V_1=0, V_3=1) \numberthis \\
P(V_4=0 | V_1=1, V_3=1) &= P(V_4=0 | V_1=1, V_2=0, V_3=1) P(V_2=0 | V_1=1, V_3=1) \\
	& \qquad  + P(V_4=0 | V_1=1, V_2=1, V_3=1) P(V_2=1 | V_1=1, V_3=1) \numberthis \\
P(V_4=0 | V_1=0, V_3=2) &= P(V_4=0 | V_1=0, V_2=0, V_3=2) P(V_2=0 | V_1=0, V_3=2) \\
	& \qquad  + P(V_4=0 | V_1=0, V_2=1, V_3=2) P(V_2=1 | V_1=0, V_3=2) \numberthis \\
P(V_4=0 | V_1=1, V_3=2) &= P(V_4=0 | V_1=1, V_2=0, V_3=2) P(V_2=0 | V_1=1, V_3=2) \\
	& \qquad  + P(V_4=0 | V_1=1, V_2=1, V_3=2) P(V_2=1 | V_1=1, V_3=2) \numberthis 
\end{align*}

\begin{align*}
P(V_4=0 | V_2=0, V_3=0) &= P(V_4=0 | V_1=0, V_2=0, V_3=0) P(V_1=0 | V_2=0, V_3=0) \\
	& \qquad  + P(V_4=0 | V_1=1, V_2=0, V_3=0) P(V_1=1 | V_2=0, V_3=0) \numberthis \\
P(V_4=0 | V_2=1, V_3=0) &= P(V_4=0 | V_1=0, V_2=1, V_3=0) P(V_1=0 | V_2=1, V_3=0) \\
	& \qquad  + P(V_4=0 | V_1=1, V_2=1, V_3=0) P(V_1=1 | V_2=1, V_3=0) \numberthis \\
P(V_4=0 | V_2=0, V_3=1) &= P(V_4=0 | V_1=0, V_2=0, V_3=1) P(V_1=0 | V_2=0, V_3=1) \\
	& \qquad  + P(V_4=0 | V_1=1, V_2=0, V_3=1) P(V_1=1 | V_2=0, V_3=1) \numberthis \\
P(V_4=0 | V_2=1, V_3=1) &= P(V_4=0 | V_1=0, V_2=1, V_3=1) P(V_1=0 | V_2=1, V_3=1) \\
	& \qquad  + P(V_4=0 | V_1=1, V_2=1, V_3=1) P(V_1=1 | V_2=1, V_3=1) \numberthis \\
P(V_4=0 | V_2=0, V_3=2) &= P(V_4=0 | V_1=0, V_2=0, V_3=2) P(V_1=0 | V_2=0, V_3=2) \\
	& \qquad  + P(V_4=0 | V_1=1, V_2=0, V_3=2) P(V_1=1 | V_2=0, V_3=2) \numberthis \\
P(V_4=0 | V_2=1, V_3=2) &= P(V_4=0 | V_1=0, V_2=1, V_3=2) P(V_1=0 | V_2=1, V_3=2) \\
	& \qquad  + P(V_4=0 | V_1=1, V_2=1, V_3=2) P(V_1=1 | V_2=1, V_3=2) \numberthis \\
P(V_4=1 | V_1=0, V_2=0) &= P(V_4=1 | V_1=0, V_2=0, V_3=0) P(V_3=0 | V_1=0, V_2=0) \\
	& \qquad + P(V_4=1 | V_1=0, V_2=0, V_3=1) P(V_3=1 | V_1=0, V_2=0) \numberthis \\
	& \qquad + P(V_4=1 | V_1=0, V_2=0, V_3=2) P(V_3=2 | V_1=0, V_2=0) \\
P(V_4=1 | V_1=1, V_2=0) &= P(V_4=1 | V_1=1, V_2=0, V_3=0) P(V_3=0 | V_1=1, V_2=0) \\
	& \qquad  + P(V_4=1 | V_1=1, V_2=0, V_3=1) P(V_3=1 | V_1=1, V_2=0) \numberthis \\
	& \qquad  + P(V_4=1 | V_1=1, V_2=0, V_3=2) P(V_3=2 | V_1=1, V_2=0) \\
P(V_4=1 | V_1=0, V_2=1) &= P(V_4=1 | V_1=0, V_2=1, V_3=0) P(V_3=0 | V_1=0, V_2=1) \\
	& \qquad  + P(V_4=1 | V_1=0, V_2=1, V_3=1) P(V_3=1 | V_1=0, V_2=1) \numberthis \\
	& \qquad  + P(V_4=1 | V_1=0, V_2=1, V_3=2) P(V_3=2 | V_1=0, V_2=1) \\
P(V_4=1 | V_1=1, V_2=1) &= P(V_4=1 | V_1=1, V_2=1, V_3=0) P(V_3=0 | V_1=1, V_2=1) \\
	& \qquad  + P(V_4=1 | V_1=1, V_2=1, V_3=1) P(V_3=1 | V_1=1, V_2=1) \numberthis \\
	& \qquad  + P(V_4=1 | V_1=1, V_2=1, V_3=2) P(V_3=2 | V_1=1, V_2=1) 
\end{align*}

\begin{align*}
P(V_4=1 | V_1=0, V_3=0) &= P(V_4=1 | V_1=0, V_2=0, V_3=0) P(V_2=0 | V_1=0, V_3=0) \\
	& \qquad  + P(V_4=1 | V_1=0, V_2=1, V_3=0) P(V_2=1 | V_1=0, V_3=0) \numberthis \\
P(V_4=1 | V_1=1, V_3=0) &= P(V_4=1 | V_1=1, V_2=0, V_3=0) P(V_2=0 | V_1=1, V_3=0) \\
	& \qquad  + P(V_4=1 | V_1=1, V_2=1, V_3=0) P(V_2=1 | V_1=1, V_3=0) \numberthis \\
P(V_4=1 | V_1=0, V_3=1) &= P(V_4=1 | V_1=0, V_2=0, V_3=1) P(V_2=0 | V_1=0, V_3=1) \\
	& \qquad  + P(V_4=1 | V_1=0, V_2=1, V_3=1) P(V_2=1 | V_1=0, V_3=1) \numberthis \\
P(V_4=1 | V_1=1, V_3=1) &= P(V_4=1 | V_1=1, V_2=0, V_3=1) P(V_2=0 | V_1=1, V_3=1) \\
	& \qquad  + P(V_4=1 | V_1=1, V_2=1, V_3=1) P(V_2=1 | V_1=1, V_3=1) \numberthis \\
P(V_4=1 | V_1=0, V_3=2) &= P(V_4=1 | V_1=0, V_2=0, V_3=2) P(V_2=0 | V_1=0, V_3=2) \\
	& \qquad  + P(V_4=1 | V_1=0, V_2=1, V_3=2) P(V_2=1 | V_1=0, V_3=2) \numberthis \\
P(V_4=1 | V_1=1, V_3=2) &= P(V_4=1 | V_1=1, V_2=0, V_3=2) P(V_2=0 | V_1=1, V_3=2) \\
	& \qquad  + P(V_4=1 | V_1=1, V_2=1, V_3=2) P(V_2=1 | V_1=1, V_3=2) \numberthis \\
P(V_4=1 | V_2=0, V_3=0) &= P(V_4=1 | V_1=0, V_2=0, V_3=0) P(V_1=0 | V_2=0, V_3=0) \\
	& \qquad  + P(V_4=1 | V_1=1, V_2=0, V_3=0) P(V_1=1 | V_2=0, V_3=0) \numberthis \\
P(V_4=1 | V_2=1, V_3=0) &= P(V_4=1 | V_1=0, V_2=1, V_3=0) P(V_1=0 | V_2=1, V_3=0) \\
	& \qquad  + P(V_4=1 | V_1=1, V_2=1, V_3=0) P(V_1=1 | V_2=1, V_3=0) \numberthis \\
P(V_4=1 | V_2=0, V_3=1) &= P(V_4=1 | V_1=0, V_2=0, V_3=1) P(V_1=0 | V_2=0, V_3=1) \\
	& \qquad  + P(V_4=1 | V_1=1, V_2=0, V_3=1) P(V_1=1 | V_2=0, V_3=1) \numberthis \\
P(V_4=1 | V_2=1, V_3=1) &= P(V_4=1 | V_1=0, V_2=1, V_3=1) P(V_1=0 | V_2=1, V_3=1) \\
	& \qquad  + P(V_4=1 | V_1=1, V_2=1, V_3=1) P(V_1=1 | V_2=1, V_3=1) \numberthis \\
P(V_4=1 | V_2=0, V_3=2) &= P(V_4=1 | V_1=0, V_2=0, V_3=2) P(V_1=0 | V_2=0, V_3=2) \\
	& \qquad  + P(V_4=1 | V_1=1, V_2=0, V_3=2) P(V_1=1 | V_2=0, V_3=2) \numberthis \\
P(V_4=1 | V_2=1, V_3=2) &= P(V_4=1 | V_1=0, V_2=1, V_3=2) P(V_1=0 | V_2=1, V_3=2) \\
	& \qquad  + P(V_4=1 | V_1=1, V_2=1, V_3=2) P(V_1=1 | V_2=1, V_3=2) \numberthis 
\end{align*}

\end{landscape}

